\documentclass[journal]{IEEEtran}
\IEEEoverridecommandlockouts
\pdfoutput=1

\usepackage[utf8]{inputenc}
\usepackage{xcolor}
\usepackage{soul}
\usepackage{amsfonts}
\usepackage{amsmath}

\usepackage[caption=false,font=footnotesize]{subfig}
\usepackage[ruled]{algorithm2e}
\usepackage{hyperref}
\usepackage{mathtools}
\usepackage[nolist]{acronym}
\usepackage[numbers]{natbib}
\usepackage{tabulary}
\usepackage{hhline}
\usepackage{graphicx}
\usepackage{dsfont}
\usepackage{array}
\usepackage{multirow}
\usepackage{numprint}
\npthousandsep{,}
\usepackage{makecell}
\usepackage[shortlabels]{enumitem}
\newcommand{\subparagraph}{}
\usepackage{titlesec}
\titlespacing*{\section}
{0pt}{1ex minus .5ex}{1ex minus 0.5ex}
\titlespacing*{\subsection}
{0pt}{1ex minus 1ex}{1ex plus .2ex minus 1ex}

\newcommand{\camera}{c}

\newcommand{\imageidx}{i}
\newcommand{\datapoint}{x}
\newcommand{\dataset}{\mathbf{X}}
\newcommand{\image}{v}

\newcommand{\labelword}{\lambda}
\newcommand{\termidx}{j}
\newcommand{\labelvec}{\ell}
\newcommand{\labelmat}{\Lambda}
\newcommand{\labelset}{\mathcal{L}}
\newcommand{\labelsignal}{\Lambda_\camera^\termidx}
\newcommand{\countimages}{n}
\newcommand{\numimages}{{N}}
\newcommand{\numimagescam}{N_c}
\newcommand{\numlabels}{{M}}
\newcommand{\numwindows}{{W}}
\newcommand{\numwindowsprime}{{W_{\refcorpus}}}
\newcommand{\timestamp}{t}
\newcommand{\topic}{z}

\newcommand{\numtopics}{{K}}
\newcommand{\numtopicsstar}{{K^*}}
\newcommand{\changeptidx}{r}
\newcommand{\numchangepts}{R}
\newcommand{\changeptvec}{\mathbf{\rho}}

\newcommand{\docfreq}{f}
\newcommand{\weight}{w}
\newcommand{\topicwordparam}{\beta}
\newcommand{\topicworddist}{\phi}
\newcommand{\doctopicparam}{\alpha}
\newcommand{\doctopicdist}{\theta}
\newcommand{\topicsignal}{\Theta_\camera^\topic}
\newcommand{\fd}[1]{D_{\text{f}}(#1)}

\newcommand{\rpd}[1]{D_{\text{RP}}(#1)}
\newcommand{\symfd}[1]{D_{\text{f}}^\text{sym}(#1)}

\newcommand{\symestrpd}[1]{\widehat{D}_{\text{RP}}^\text{sym}(#1)}

\makeatletter
\newcommand{\distas}[1]{\mathbin{\overset{#1}{\kern\z@\sim}}}%
\newcommand{\boldx}{\mathbf{\omega}}

\newcommand{\refcorpus}{\mathbf{Y}}

\newcommand{\thresh}{\tau}

\begin{acronym}
\acro{API}{application programming interface}
\acro{CNN}{convolutional neural network}
\acro{GCV}[GCV]{Google Cloud Vision}
\acro{LDA}[LDA]{Latent Dirichlet Allocation}
\acro{LSI}[LSI]{Latent Semantic Indexing}
\acro{SVD}[SVD]{Singular Value Decomposition}
\acro{NLP}[NLP]{Natural Language Processing}
\acro{tf-idf}{Term Frequency-Inverse Document Frequency}
\acro{MassDOT}{Massachusetts Department of Transportation}
\acro{BoW}{Bag-of-Words}
\acroplural{BoWs}{Bags-of-Words}
\acro{BoLW}{Bag-of-Label-Words}
\acroplural{BoLW}[BoLWs]{Bags-of-Label-Words}
\acro{BoVW}{Bag-of-Visual-Words}
\acro{NOAA-GHCN}{NOAA Global Historical Climatology Network}
\acro{GPU}{Graphics Processing Unit}
\acro{uLSIF}{unconstrained Least-Squares Importance Fitting}
\acro{RuLSIF}{Relative unconstrained Least-Squares Importance Fitting}
\acro{KLIEP}{Kullback--Leibler Importance Estimation Procedure}
\acro{SVD}{Singular Value Decomposition}
\acro{RPDAS}{Relative Pearson Divergence Anomaly Score}
\acro{t-SNE}{t-distributed Stochastic Neighbor Embedding}
\acro{PR}{Precision-Recall}
\acro{TMC}{Traffic Management Center}
\acro{CCTV}{Closed-Circuit Television}
\acro{ILSVRC}{ImageNet Large Scale Visual Recognition Challenge}
\acro{BFCC}{Boston Freeway CCTV Camera}
\end{acronym}

\begin{document}
\title{Semantic Analysis of Traffic Camera Data: Topic Signal Extraction and Anomalous Event Detection
\thanks{This work was performed under the financial assistance award PSIAP3774 from U.S. Dept. of Commerce, National Institute of Standards and Technology}
\thanks{We also acknowledge support from National Science Foundation grants CNS-1239054 and CNS-1453126, and FM IRG within the Singapore-MIT Alliance for Research and Technology.}
}

\author{Jeffrey~Liu\IEEEauthorrefmark{1}\IEEEauthorrefmark{2},
        Andrew~Weinert\IEEEauthorrefmark{2},
        and~Saurabh~Amin\IEEEauthorrefmark{1}
\thanks{\IEEEauthorrefmark{1}J. Liu and S. Amin were with the Department
of Civil and Environmental Engineering, Massachusetts Institute of Technology, Cambridge,
MA, 02130 USA, e-mail: jeffliu@mit.edu, amins@mit.edu.}
\thanks{\IEEEauthorrefmark{2}J. Liu and A. Weinert were with Massachusetts Institute of Technology, 
Lincoln Laboratory, Lexington, MA 02421, USA, email: jeffrey.liu@ll.mit.edu, andrew.weinert@ll.mit.edu.}}
\maketitle
\begin{abstract}
\acfp{TMC} routinely use traffic cameras to provide situational awareness regarding traffic, road, and weather conditions. Camera footage is quite useful for a variety of diagnostic purposes; yet, most footage is kept for only a few days, if at all. This is largely due to the fact that currently, identification of notable footage is done via manual review by human operators---a laborious and inefficient process. In this article, we propose a semantics-oriented approach to analyzing sequential image data, and demonstrate its application for automatic detection of real-world, anomalous events in weather and traffic conditions. Our approach constructs semantic vector representations of image contents from textual labels which can be easily obtained from off-the-shelf, pretrained image labeling software. These semantic label vectors are used to construct \emph{semantic topic signals}---time series representations of physical processes---using the \acf{LDA} topic model. By detecting anomalies in the topic signals, we identify notable footage corresponding to winter storms and anomalous traffic congestion. In validation against real-world events, anomaly detection using semantic topic signals significantly outperforms detection using any individual label signal. 
\end{abstract}

\pdfoutput=1
\section{Introduction}
\label{sec:introduction}
\subsection{Motivation}
\acf{CCTV} traffic cameras are common sensors used by many \acfp{TMC} to provide situational awareness of road infrastructure networks. Cameras provide rich, intuitive, visual information about driving conditions, infrastructure health, and traffic congestion 24 hours a day. It is interesting to note that most of this footage is kept only for a few hours or days, if at all, before being permanently deleted \cite{Kuciemba2016}. In some ways, it is sensible not to store all of the footage: most of the time, nothing out of the ordinary is happening, and video requires large amounts of disk storage. Yet, discarding all of this data is also a potential waste of rich data from a widely-deployed, flexible sensor, which could be used to improve traffic analytics or diagnostics of infrastructure performance. 

Indeed, a small amount of ``notable'' footage does get manually saved by operators for training personnel, performing diagnostics, or providing documentation \cite{Kuciemba2016}. However, this process is inefficient and potentially inconsistent: humans struggle to parse video information from more than one source at a time \cite{Marois2005}, and most \acp{TMC} have hundreds of cameras which run 24 hours a day. It is thus infeasible for human operators to constantly monitor all of the incoming footage simultaneously. Furthermore, only a few \acp{TMC} in the US have written policies regarding what footage is notable enough to be saved \cite{Kuciemba2016}. Even for those \acp{TMC} that do, the policy's execution is subject to human factors such as subjective interpretation, fatigue, and distraction. In this article, we seek to address the problem of automatically and consistently identifying notable events from sequential image data, particularly traffic \ac{CCTV} footage. 

Toward addressing this challenge, we develop a \acf{NLP}-inspired, methodological approach to analyzing sequential image data, which seeks to preserve the intuitive and human-interpretable nature of images. As a starting point, image contents are represented as \acf{BoLW} semantic feature vectors constructed from labels from off-the-shelf image labeling software. These semantic feature vectors are used in the \acf{LDA} topic model to infer \emph{semantic topic signals}---time series corresponding to physical processes shown in the footage, such as winter storms and traffic congestion. The semantic topic signals are then analyzed to identify notable events corresponding to changes and anomalies in the signals. In particular, we employ a direct divergence estimation technique based on \cite{Yamada2013} for anomaly detection which does not require parametrically fitting the test and reference data distributions. Furthermore, we present a new, public dataset of real-world traffic camera footage, which serves as the basis for the empirical demonstration and evaluation of our approach.

\subsection{Contributions and Prior Literature}
We now discuss this article's contributions, the associated article sections, and the relevant prior literature: 

\textbf{1.) \acl{BFCC} Dataset.} In Sec.~\ref{sec:case_study}, we introduce our \ac{BFCC} dataset containing \numprint{259,830} frames of traffic \ac{CCTV} footage from Boston-area freeway cameras, annotated with a broad vocabulary of labels using commercial image labeling services. 

Public traffic \ac{CCTV} datasets relatively recent additions to the transportation literature. We identified two previously published traffic \ac{CCTV} datasets: WebCamT \cite{Zhang2017a} and the Car Accident Detection and Prediction (CADP) dataset \cite{Shah2018CADPAN}. WebCamT claimed to be the first publicly available dataset of traffic camera footage \cite{Zhang2017a}. It provides detailed annotations for the footage: bounding boxes and labels for vehicle types and weather, as well as vehicle counts and re-identification \cite{Zhang2017a}. The data in WebCamT were collected over four separate, one-hour periods for each day, at a sample rate of one frame per second. The second dataset, CADP, collects and annotates video segments of vehicle crashes from YouTube with vehicle bounding boxes \cite{Shah2018CADPAN}. The videos in CADP are short (a few minutes on average) and intermittent, since they only include crashes. In comparsion, the \ac{BFCC} provides continuous 24-hour footage from the traffic cameras.

The image labels for the \ac{BFCC} dataset are generated from a pretrained image labeling service powered by deep learning models, such as \acfp{CNN}. In recent years, the performance of image labeling algorithms have improved to human-comparable error rates in the \ac{ILSVRC} benchmark \cite{ILSVRC15}. Though deep learning approaches achieve remarkable performance, they are much more computationally expensive and data-intensive to train than classical approaches \cite{Livni2014a}. Consequently, many developers and organizations now offer free \cite{Tensorflow2018} and commercial \cite{GoogleImageBeta} pretrained, off-the-shelf image labeling tools and services to detect a wide array of object classes. 

\ac{CNN}-based techniques have been used to recognize traffic congestion directly from images. For example, \cite{Pamula2018} trains a \ac{CNN} to recognize different levels of congestion as classes; \cite{Luo2018} trains a classifier to segment the image between road and vehicle and compute the density directly; and \cite{Zhang2017a} takes a similar approach, but also estimates a density map to correct for the distortion effects of perspective and distance in the image. These approaches are indeed performant, but require large amounts of data and computation to train \cite{Livni2014a}. In comparison, our approach leverages general-purpose, pretrained image labeling software, and is able to detect multiple processes, such as traffic congestion and weather, without needing to build and train custom, bespoke models. However, we acknowledge that our approach is best suited for detecting qualitative differences in processes, and is not meant to be a precise estimator of quantities such as vehicle density.

\textbf{2.) \acf{BoLW} Model and Semantic Features.} We present the \ac{BoLW} model in Sec.~\ref{sec:bolw}, based off the well-known \ac{NLP} \acf{BoW} model \cite{harris1954distributional}, for representing the image contents as \emph{semantic feature vectors}. There is a related \ac{BoW} model in computer vision called \acf{BoVW} \cite{Li2005}, where the \emph{visual words} are visual features, such as pixel clusters. In contrast, our ``words'' are semantic labels---literal textual words. The \ac{BoW} and \ac{BoVW} models serve as foundational models for a variety of techniques in \ac{NLP} and computer vision because they enable linear algebra to be performed on documents and visual contents of images respectively \cite{Manning2009a,Li2005}. Our \ac{BoLW} model seeks to accomplish the same for the semantic contents of images.

\textbf{3.) Identification of Topic Signals via \ac{LDA}.} We present an approach to inferring semantic \emph{topic signals} via the \ac{LDA} topic model in Sec.~\ref{sec:topic_models}. Topics are distributions of labels, and can correspond to physical processes such as storms and traffic congestion. Topic signals represent the fraction of semantic contents related to the given topic over time. \ac{LDA} is an \ac{NLP} topic model \cite{Blei2003a,Griffiths2004}, which is used to find topics---distributions of related words---in a corpus of documents, and characterize the documents in terms of these topics. Previous works in the transportation literature have used topic models to analyze written documents, such as failure reports for railway \cite{Wang2017} and aviation applications \cite{Kuhn2018}. Additionally, there has has been work in applying \ac{LDA} to \emph{visual word} features for dimensionality reduction of image and video data \cite{Li2005, Roller2013}. However, aside from our earlier conference publication \cite{Liu2018a}, we are not aware of any prior use of \ac{LDA} to identify topics and signals from semantic labels of sequential image data. 

\textbf{4.) Detection of Notable Footage from Topic Signals.} Using the semantic topic signals, we formulate the detection of notable footage as change and anomaly detection problems in the topic signal in Sec.~\ref{sec:anomaly_detection}. Anomaly detection is concerned with identifying data that are unlikely to come from a reference distribution \cite{Hodge2004}. Change detection is a special instance of anomaly detection, where the reference distribution is given by the data in the immediately-preceding time interval.

Change detection is well studied in image processing contexts \cite{Tewkesbury2015}; however, change detection in image processing is typically concerned with detection of changes in visual elements, such as in shapes, colors, or textures. In contrast, our approach considers changes in the semantic representations of image contents. Where there exist some prior work in using semantic information to contextualize detected visual changes \cite{Suzuki2016}, we could not find any examples of applications which considered purely-semantic representations of image data.

Our anomaly detection procedure uses divergence measures to quantify the dissimilarity between the test and reference distributions of data. We utilize a technique which allows us to directly compute the divergence without needing to estimate parametric forms of the respective distributions. This is based on the \ac{RuLSIF} procedure \cite{Yamada2013}, which is derived from the \ac{uLSIF} procedure \cite{Kanamori2009}, and the \ac{KLIEP} \cite{Sugiyama2007}. Direct estimation techniques have been applied to outlier and change point detection generally in \cite{Hido2011, Liu2013b}, but our use of such techniques in detecting anomalies in semantic representations of sequential image data is novel.

\textbf{5.) Empirical Evaluation}
We provide empirical results and evaluation of the performance of the aforementioned techniques on the \ac{BFCC} dataset. These results are validated against known disruption events---including holidays, city parking bans, and winter storms---and are discussed throughout the article in the sections corresponding to the respective methods. The empirical results serve as proofs-of-concept and demonstrations of our methods and approach. 

This article is an extension of our earlier conference publication \cite{Liu2018a}. In order to paint a complete picture, some elements have been included from the conference version. However, this article offers significantly more results and methods, particularly in Sec.~\ref{sec:topic_models} and the entirety of Sec.~\ref{sec:anomaly_detection}.

\section{Traffic Camera Data}
\label{sec:case_study}
In this section we give an overview of the \acf{BFCC} dataset of traffic \ac{CCTV} images and semantic labels, which serves as the basis for empirical results presented in this paper. 

While there exists other traffic camera datasets, such as WebCamT \cite{Zhang2017a} and CADP \cite{Shah2018CADPAN}, the \ac{BFCC} dataset seeks to provide a greater breadth of labels and time periods covered. We offer scene-level annotations for several hundred label classes generated by a commercial image labeling service, which serve as semantic label features, as well as performance benchmarks of a general-purpose commercial image labeling service. In addition, footage in the \ac{BFCC} dataset covers all 24 hours of the day, instead of just a few select hours or events. 

\subsection{CCTV Footage}
\begin{figure}[htb]
\centering
\includegraphics[width=0.6\linewidth]{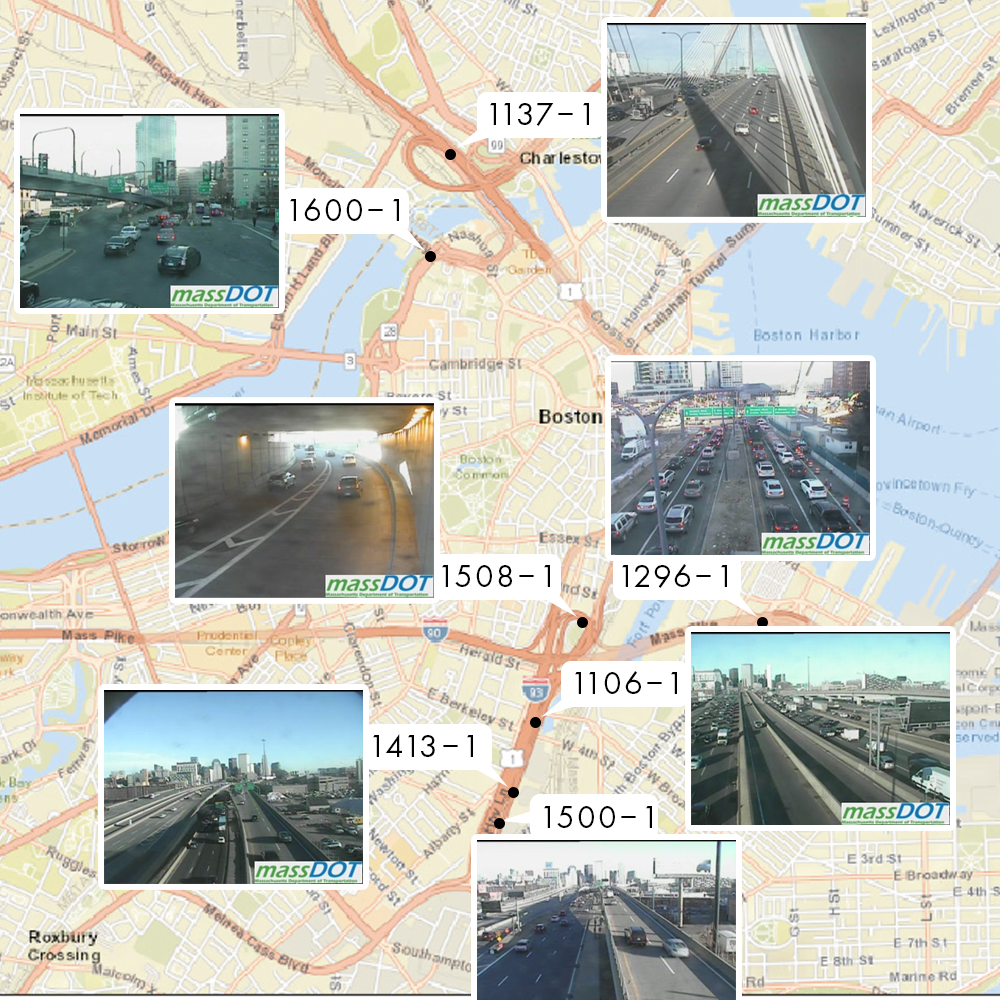}%
\caption{Camera locations and sample images. We selected a diverse set of cameras which depicted several different network locations and components, including a bridge (1137--1), underpass (1508--1), intersection (1600--1), HOV lane (1106--1), median (1296--1), and open freeway (1413--1, 1500--1)}
\label{fig:cam_details}
\end{figure}

We collected footage from seven \ac{MassDOT} freeway \ac{CCTV} cameras in the Boston metro area. The footage was obtained by scraping the public Mass511 Traveler Information Service website and saved as individual frames (also referred to generically as \emph{images}). Each frame has a resolution of $320 \times 240$ pixels, and there is a sampling period of roughly 3 minutes between frames, which represents the lower end of camera resolution and frame rate capabilities for typical traffic \acp{CCTV} \cite{Kuciemba2016}. These cameras were used in our earlier conference paper \cite{Liu2018a}, which also includes additional details about each camera.

Fig.~\ref{fig:cam_details} provides additional details about each camera, including their respective locations, \ac{MassDOT}-assigned identification numbers and names, and sample images. Each camera is remotely controllable with by the \ac{MassDOT} \ac{TMC} operators, with pan/tilt/zoom capabilities. Some of the cameras in the dataset were frequently repositioned to view alternate perspectives, focus on specific areas of the road, or to avoid obstruction due to snow accumulation. 

The data were collected in two phases: phase I consists of the week November 6$^\text{th}$--November 12$^\text{th}$, 2017; phase II consists of the period between December 17$^\text{th}$, 2017--January 31$^\text{st}$, 2018. A total of 259830 frames were collected over these two periods. Phase I data serves as an experimental baseline ``reference" dataset which was used to establish the road network's behavior under nominal conditions. Phase I contained no storms or precipitation, but did include the Veteran's day holiday on Saturday, November 11$^\text{th}$ (observed on the $10^\text{th}$). The data from phase II serves as the experimental ``test" dataset. Notable events that occurred during this phase include: the Christmas and New Year's holidays; several snow storms, including the ``bomb cyclone" winter storm of January 2018; and a two-day parking ban imposed by the City of Boston in response to the ``bomb cyclone" storm. 

Table~\ref{tab:anomalies} lists the notable events we considered for validation tasks in Section~\ref{sec:anomaly_detection}. We considered all snowfall or rainfall events with at least 0.5" of precipitation within a 24 hour period, as reported by the \ac{NOAA-GHCN} daily records \cite{menne2012global}, as ``notable.'' For holidays and events, we considered the city-imposed parking bans \cite{Andersen} and ``major" holidays where retail stores had significantly different hours or were closed \cite{MassLive2017b,MassLive2018}. We used the criteria of modified retail hours rather than the federal holiday calendar because not all federal holidays are widely observed, and businesses generally adjust their hours in response to consumer demand. Thus, modified hours are more likely to indicate whether a holiday is widely observed. For this reason, we did not include Veteran's day or Martin Luther King Jr. day as major holidays \cite{MassLive2017,MassLive2018b}. Additionally, we did not consider Christmas Eve or New Year's Eve as major events, since they fell on Sundays, and stores in the Boston area observed their typical hours on those dates \cite{MassLive2017b,MassLive2018}. 

\begin{table}[h]
    \centering
\footnotesize\begin{tabular}{|l|l|c|c|}
        \hline
        \textbf{Date}                          & \textbf{Holiday/Event}           & \textbf{Rain} $>0.5"$ & \textbf{Snow} $>0.5"$               \\ \hline
        Dec 23, 2017    &  & \checkmark   &            \\ \hline
         Dec 25, 2017   &Christmas  &  & \checkmark              \\ \hline
         Jan 1,  2018   &New Year  &  &                   \\\hline
         Jan 4,  2018   &Parking Ban  & \checkmark   & \checkmark  \\\hline
         Jan 5,  2018   &Parking Ban  &  &  \\\hline
         Jan 12, 2018   &  &\checkmark   &                              \\\hline
         Jan 13, 2018   &  &\checkmark   &                            \\\hline
         Jan 17, 2018   &  &  & \checkmark                               \\\hline
         Jan 23, 2018   &  &\checkmark   &                             \\\hline
         Jan 30, 2018   &  &  & \checkmark      \\\hline
        \end{tabular}
        \vspace{0.2em}

    \caption{List of notable events}
    \label{tab:anomalies}
\end{table}

\subsection{Semantic Feature Labels}
\label{sub:labels}
We tag each frame of traffic \ac{CCTV} footage with \emph{labels} of the image contents using a pretrained, commercially available, common image labeling service: \acf{GCV} \cite{GoogleImageBeta}. \ac{GCV} offers a number of products, of which we use two:
the \ac{GCV} ``label detection" service, referred to as Label Source 1 (LS1), and the \ac{GCV} ``web entity detection" service (resp. LS2). LS1 provides annotations for ``broad sets of categories within an image, ranging from modes of transportation to animals," \cite{GoogleClientAPIDoc}, while LS2 integrates additional information and metadata from the web, such as links and related websites, to detect ``web entities"---web searches related to the image \cite{GoogleClientAPIDoc}. 

Note that our techniques are not exclusive to the \ac{GCV} services, and can be applied using any image labeling implementation. However, our techniques do assume that the image content recognition problem is a \emph{multi-label} classification problem, where each image can be tagged with multiple labels, as opposed to a \emph{multi-class} problem, where each image is classified into exactly one class \cite{zhang_review_2014}. This is because we consider the distribution of labels and their co-occurrence to extract semantic topic signals, which necessitates multiple labels per image.

We chose the \ac{GCV} commercial implementation because it covers a broad set of categories, is actively maintained and documented, and required less technical overhead for the user compared to open-source, locally-deployed solutions. In terms of breadth of categories, \ac{GCV} included labels corresponding to ``traffic" and ``traffic congestion" in both LS1 and LS2. None of the open source implementations that we examined---including ImageNet \cite{ILSVRC15}, and Places365 \cite{zhou2017places}---included such labels in their classification set. We use these labels as benchmarks for comparison for the performance of our ``Traffic congestion" \emph{topic signal} in Sections~\ref{sec:topic_models} and \ref{sec:anomaly_detection}. 

In terms of convenience and technical overhead, the commercial implementations required less effort to set up than the open-source ones. The commercial implementations operate as cloud services \cite{GoogleClientAPIDoc}, where the user submits an HTTP POST request with the image, and receives a list of labels in return. This can be done independent of programming language or operating system. In comparison, most pretrained open source implementations required the user to install specific libraries and frameworks in order to run. While this is still easier than training an image labeling model from scratch, it imposes additional technical overhead to the user. For prototyping purposes, the commercial implementations allow for quick annotation of images and the identification of application-relevant labels.

\begin{figure}[h]
\centering
\subfloat[Camera 1137--1, 2018-01-04 16:57:52 (UTC) ]{\raisebox{-0.5in}{\includegraphics[width=0.4\linewidth]{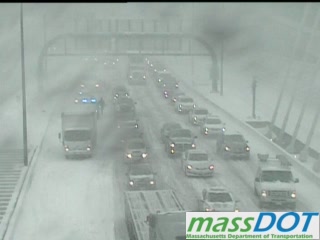}}%
\label{fig:annotation_example}}
\subfloat[Labels for sample image]{
\footnotesize
\begin{tabular}{|m{0.25\linewidth}|}
\hline
    \textbf{LS1 labels} \\ \hline
        snow \\\hline
    	infrastructure \\\hline
    	mode of transport \\\hline
    	lane\\\hline
    	winter storm \\\hline
    	road\\\hline
    	transport\\\hline
    	structure\\\hline
    	phenomenon\\\hline
    	blizzard\\\hline
    	highway\\\hline
    	freezing\\\hline
    	automotive exterior\\\hline
    	glass\\\hline
\end{tabular}
\begin{tabular}{|m{0.2\linewidth}|}
    \hline
     \textbf{LS2 labels}\\ \hline
        Blizzard \\\hline
        Lane\\\hline
        Car\\\hline
        Transport \\\hline
        Snow \\\hline
        Highway	\\\hline
        Fog \\\hline
        Glass \\\hline
        Freezing \\\hline
        \makecell[bl]{Massachusetts\\ Department of\\ Transportation} \\ \hline
    \end{tabular}
\label{tab:labels}}
\caption{Fig.\protect\subref{fig:annotation_example} shows a sample image taken during the ``bomb cyclone," and the table in \protect\subref{tab:labels} shows the labels returned by each labeling service}
\label{fig:label_samples}
\end{figure}

Fig.~\ref{fig:label_samples} presents the labels reported by each label service for a sample image taken from Camera 1137--1 during the ``bomb cyclone'' blizzard. We refer to the set of all possible labels from the label services as the \emph{vocabulary}. Some labels appear in both services, but not necessarily on the same images. For example, ``road'' appears in both label sources, but for example in Fig.~\ref{fig:label_samples}, it is only reported by LS1. Thus, to disambiguate between the labels from each source, we prepend all label text with the respective label source identifier, e.g. ``LS1: snow" vs. ``LS2: Snow".\footnote{In addition, labels from LS1 were reported by the service in lowercase, whereas those from LS2 were rendered with capitalizations. We preserve this styling.} This convention would allow additional label sources to be incorporated without ambiguity in future work by prepending the respective labels with ``LS3:'', ``LS4'', etc.
In this article, if we refer to a label generically without its label source identifier (e.g. the label ``snow'') we are referring to \emph{both} of the labels from each source (i.e. ``LS1: snow'' and ``LS2: Snow''). 

The size of the vocabulary for the labels in the dataset is 1389 total labels: 477 from LS1, and 912 from LS2. 
In general, the labels from LS2 tend to be more specific and contain more named entities than those from LS1. For example, we observed the labels ``LS2: BMW," ``LS2: BMW 3 Series," and ``LS2: 2018 BMW 3 Series Sedan" from LS2, whereas we found only the label ``LS1: bmw" from LS1. However, LS2 was also prone to including more spurious labels due to word associations: for example, the label ``LS2: Blizzard Entertainment" (a software company), appeared occasionally alongside ``LS2: Blizzard."  Fortunately, such spurious labels were rare, and we were able to address this issue in our analysis with a high-pass filter on the labels' empirical document frequency $\docfreq^\termidx$, given by
   $ \docfreq^\termidx := {\countimages^\termidx/\numimages}$
where $\countimages^\termidx$ is the number of images in the dataset in which the label $\termidx$ appears, and $\numimages$ is the total number of images in the dataset. 

The cutoff for the high-pass filter is set at $\docfreq^\termidx = 10^{-4}$, and was chosen heuristically. We considered that spurious labels may show up once or twice per camera; thus, we set the cutoff at a baseline average rate of three images per camera, which corresponds to a fraction of roughly $0.01\%$ all frames. We also verified that the remaining labels were related to objects and phenomena likely to be observed in traffic footage. In addition, we removed labels from our analysis related to ``Massachusetts Department of Transportation," as those labels are likely due to the ``massDOT" watermark in the corner of each image, and not the scene content. After filtering, we were left with 620 labels in the vocabulary: 280 from LS1 and 340 from LS2.

\section{Bag of Label Words}
\label{sec:bolw}


\subsection{Model}
Consider a set of $\numimages$ images (frames of traffic \ac{CCTV} footage\footnote{We use the generic term \emph{image} instead of \emph{frame} when discussing the \ac{BoLW} and \ac{LDA} models in this article, as they are applicable to any collection of images.}), indexed by $\imageidx \in [1, \dots, \numimages]$. Each image has an originating camera, denoted $\camera_\imageidx$, and timestamp, denoted $\timestamp_\imageidx$. The number of images from a given camera is denoted $\numimagescam$.

We now present the \ac{BoLW} model for representing the image contents as a semantic feature vector. Consider a vector space, $\labelset$, where each dimension corresponds to an individual label in the label vocabulary. The dimension of $\labelset$---the number of terms in the label vocabulary---is denoted $\numlabels$. A vector in this vector space $\labelvec \in \labelset$ represents the labels of image $\imageidx$. The nonzero entries of $\labelvec_\imageidx$ are equal to unity in the dimensions corresponding to each of the semantic labels for image $\imageidx$. This vector representation is analogous to the \acl{BoW} vector space model of documents in \ac{NLP}, which represents documents as vectors, where each component corresponds to the number of occurrences of a given word in the document \cite{Manning2009a}; hence, we refer to our model as \acl{BoLW}. 

The \acl{BoLW} vector model is given as follows:
\begin{itemize}
    \item A \emph{label word}, $\labelword_\termidx$, is defined as a single label in the label vocabulary, indexed by $\termidx \in \{1, \dots, \numlabels\}$. $\labelword_\termidx$ is a (one-hot) unit-basis vector in $\labelset$ whose $\termidx^\text{th}$ component equals one, and all other components equal zero.
    \item A \emph{bag of label words} associated with image $\imageidx$ is a vector $\labelvec_\imageidx \in \labelset$. 
    \item The total \emph{weight}, $\weight_\imageidx$, of bag $\labelvec_\imageidx$ is defined as its $L^1$-norm: $\weight_\imageidx := \Vert \labelvec_\imageidx \Vert_1 = \sum_\termidx |\labelvec_\imageidx^\termidx|$
\end{itemize}

There is another related \ac{BoW} model in computer vision, called Bag-of-Visual-Words (BoVW) \cite{Yang2007}; however, BoVW uses pixel groupings as its ``words," whereas \ac{BoLW} uses textual, semantic labels as its ``words." Generically, these types of ``Bag-of-Words''-style models are referred to as ``Bag-of-Features'' models.
In all Bag-of-Features models, the absolute configuration of the features---word order in \ac{BoW}, pixel clusters locations in \ac{BoVW}, and labeled object positions in \ac{BoLW}---is ignored. Instead, the vector representation retains information about the presence and \emph{co-occurrence} of features. This provides invariance to certain transformations of the original data, such as permutations in word order for \ac{BoW} or rearranging of image elements in \ac{BoVW} and \ac{BoLW}.

Using \ac{BoLW}, the semantic content of the footage can be represented in a conventional matrix format. Vertically concatenating the row vectors $\labelvec_\imageidx$, ordered by timestamp, for all images of a given camera, generates the $\numimagescam \times \numlabels$ \emph{image-label} matrix $\labelmat_\camera$. Each row of the image-label matrix corresponds to an image, and each column correspond to a label.\footnote{The \emph{image-label} matrix is analogous to the \emph{document-term} matrix in \ac{NLP}, and in general, our usage of the terms ``image" and ``label" in this article correspond to ``document" and ``term" respectively in the \ac{NLP} literature.} 
This resembles a measurement matrix from signal processing: a matrix of $\numimagescam$ observations of an $\numlabels$-dimensional system. The $\termidx^\text{th}$ column of $\labelmat_\camera$ represents a (potentially unevenly-spaced) time series for label $\termidx$ and camera $\camera$. This is referred to as a \emph{label signal}, and is given by:
\begin{align}
  \Lambda^\termidx_\camera = \{\labelvec_\imageidx^\termidx; \forall \imageidx \text{ where } \camera_\imageidx = \camera \}
  \label{eq:label_signal}.
\end{align} 
If the sampling interval between images is uneven, we convert the unevenly spaced time series into a regular time series by interpolation and resampling. In our empirical analysis, we resample the data at 5-minute sampling intervals using linear interpolation.


\subsection{Label Reweighting}
\label{sub:rescale}
Note that extremely common labels, such as background elements, do not necessarily contribute much operationally useful information about the image contents. For example, labels such as ``Road" and ``Asphalt" appear extremely frequently in images in the \ac{BFCC} dataset. While these labels are not incorrect---the images from freeway cameras do indeed contain roads made of asphalt---they are also not particularly informative for \ac{TMC} operations, as it is expected that most images from a traffic camera contain a road. Thus, we would like to attenuate the weight of labels which occur extremely frequently. This is addressed with the \acf{tf-idf} weighting scheme, which rescales each image's label weights based on each label's rarity for each camera.

The \ac{tf-idf} weighting scheme is a heuristic used in \ac{NLP} to reweight terms in the \ac{BoW} vector to account for the natural difference in term prevalence in a language \cite{Manning2009a}. Terms\footnote{While in the rest of the article we use the terms ``label'' and ``image'' instead of ``term'' and ``document,'' we preserve the use of ``term'' and ``document'' in the explanation of tf-idf in this section due to those words being integral to the tf-idf (\emph{term} frequency-inverse \emph{document} frequency) name.} that are commonly used in a language will be highly represented in any given document, regardless of their relevance to the subject matter of the document. These extremely common terms can end up dominating the weight of a Bag-of-Features if all terms are weighted evenly. Thus, to correct for the effect of these prevalent terms, their weights are scaled inversely to their preponderance across all documents. Analogously, labels that appear on nearly every image tend to correspond to static background elements, such as the road and surrounding infrastructure; thus, the same \ac{tf-idf} reweighting can be used to attenuate these prevalent labels.

The \ac{tf-idf} weight is computed as the product of its two titular components: the term frequency (tf) and the inverse document frequency (idf) \cite{Manning2009a}. In \ac{NLP} usage, the term frequency of a given document and term is given by the number of occurrences of that term within the document; in our case, the term frequency for a given image $\image$ and label $\termidx$ is given by the binary variable:
\begin{align}
    \text{tf}(\imageidx, \termidx) = \begin{cases}
    1 & \text{if image $\imageidx$ has label $\termidx$}\\
    0 & \text{otherwise}
    \end{cases}.\label{eq:tf_def}
\end{align}
We use a binary tf term, since only consider the presence/absence of labels. However, the term frequency could be used more generally to represent other measures such as object count or number of pixels, if that information is available. This is beyond the scope of this article, but represents a promising refinement for future work. 

The inverse document frequency (idf) of a term $\termidx$ is typically computed as the negative logarithm the empirical document frequency: $ \text{idf}(\termidx) = -\log(\docfreq^\termidx) = \log\left(\frac{\numimages}{\countimages^\termidx}\right)$. We use a variant of idf, which we call the \emph{per-camera idf}, which is computed for camera $c$ as
$    \widetilde{\text{idf}}(\termidx, c) = \log\left(\frac{\numimagescam}{\countimages^\termidx}\right)$
where $\numimagescam$ is the total number of images for camera $c$. The per-camera idf considers the relative rarity of a label $\termidx$ within the context of the other images from that camera. This is motivated by the fact that the label distributions are different across cameras; for example, the presence of the label ``Snow" is more unusual and notable for images from a camera in a tunnel than those from a camera out in the open.

The \emph{idf-weighted image-label} matrix is a rescaling of the image-label matrix where the components of each row ($\labelvec_\imageidx$) are given by the per-camera \ac{tf-idf} values:
\begin{align}
  \labelvec_\imageidx^\termidx = \text{tf}(\imageidx, \termidx) \times \widetilde{\text{idf}}\left(\termidx, \camera(\image_\imageidx)\right). \label{eq:per_cam_tfidf}
\end{align}
The empirical analyses presented in the remainder of this article use the per-camera idf-weighted label data.

\section{Semantic Topic Signals}
\label{sec:topic_models}
In this section, we discuss the process of extracting \emph{semantic topic signals} from the \ac{BoLW} representations of sequential image data. A \emph{topic} represents a distribution of related labels, and can correspond to certain processes or phenomena, such as weather and traffic. A \emph{semantic topic signal} for a given topic and camera represents, as a function of time, the fraction of the footage's semantic contents related to that topic. 

The motivation to model processes as topics is as follows. First, certain phenomena can be modeled as random processes which generate a mix of objects (and correspondingly, labels) over time. For example, ``traffic'' can be seen as a random process which generates cars, trucks, buses, (and their respective labels) etc. at different rates. Weather can be thought of as a random process which generates rain and snow at different rates. Similarly, one can construct processes that for diurnal lighting cycles and background infrastructure. These processes can thus be modeled as probability distributions over the objects and labels that they generate.

Each frame of footage from a camera can be viewed as an observation of a mix of the aforementioned processes. Given a sufficient number of observations, one may infer the processes and their respective object generation rates, and construct signals to to represent the prevalence of those processes in the footage. We recognize that this is equivalent to the Bayesian inference problem addressed by probabilistic \emph{topic modeling} in \ac{NLP}. In particular, we use a common variant \cite{Griffiths2004} of the \acf{LDA} topic model \cite{Blei2003a}.


\subsection{Latent Dirichlet Allocation Topic Model}
\label{sub:LDA_model}

\begin{figure}[h]
    \centering
    \includegraphics[width=0.5\linewidth]{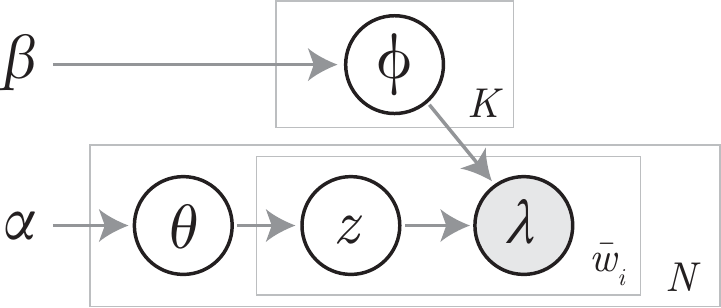}
    \caption{Graphical representation of the \ac{LDA} model structure. Each of the boxes (plates) represent a repeated component; the variable in the lower right hand corner of each plate indicates the number of copies. The outer plates represent each bag of label words in the dataset, and the inner plate represents each label word added to the bag. Grey-filled circles represent observed variables, whereas white-filled circles represent latent variables.}
    \label{fig:lda_plate}
\end{figure}

\acl{LDA} is a hierarchical Bayesian topic model for document generation in \ac{NLP}. \ac{LDA} represents documents as random mixtures of topics, denoted $\doctopicdist$, where each topic is, in turn, a probability distribution over label words, denoted $\topicworddist$. 
We originally presented the use of this model for analyzing traffic camera images in \cite{Liu2018a}. We provide a high-level overview of the \ac{LDA} model here, but refer the reader to \cite{Liu2018a} for additional details about the model. The structure of the model is illustrated in Fig.~\ref{fig:lda_plate}

%

A topic is denoted $\topic \in \{1,\dots,\numtopics\}$, where $\numtopics$ is the total number of topics, set exogenously. The \emph{topic-label} distribution is denoted $\topicworddist$, characterizes the probability distribution over labels which constitute each topic, and is drawn from a Dirichlet distribution characterized by $\numlabels$-dimensional hyperparameter $\topicwordparam$, where $\numlabels$ refers to the number of labels. $\topicworddist^{\topic} = \topicworddist(\labelword|\topic)$ denotes the label distribution for a given topic $\topic$. 

The \emph{image-topic} distribution $\doctopicdist$, represents each image as a probability distribution over topics, and is characterized by the $\numtopics$-dimensional hyperparameter $\doctopicparam$. The conditional distribution for a given topic $\topic$ is denoted $\doctopicdist^\topic(\imageidx) = \doctopicdist(\imageidx|\topic)$. 

We define the \emph{semantic topic signal} of a given topic $\topic$ and camera $\camera$ as the (potentially unevenly spaced\footnote{As with the label signals, the unevenly spaced time series are converted to regular time series through interpolation and resampling. For the empirical analysis, we resample the topic signal at 5 minute intervals with linear interpolation.}) time series: 
\begin{align}
  \topicsignal := \{\doctopicdist^\topic(\imageidx); \forall \imageidx \text{ where } \camera_\imageidx = \camera\} 
  \label{eq:topic_signal}.
\end{align} 
The topic signal represents the proportion of the camera footage's semantic weight which corresponds to topic $\topic$ over time. Increases/decreases in this signal correspond to a respective increase/decrease in the fraction of labels related to the topic. Each individual topic signal can be analyzed as a univariate time series, and combinations of topic signals can be analyzed jointly.


To fit the model, we want to find the most likely (i.e. maximum posterior probability) values for the \emph{image-topic} distribution, $\doctopicdist$, and \emph{topic-label} distribution, $\topicworddist$, given the hyperparameters and . This is done using the online variational Bayes algorithm presented in \cite{Hoffman}. We assume symmetric priors on $\doctopicdist$ and $\topicworddist$ with constant hyperparameter values $\doctopicparam={50}/{\numtopics}$ and $\topicwordparam=0.1$ based on \cite{Griffiths2004}.

\subsection{Selected Topics and Signals}
\label{sub:label_distributions}
\begin{figure*}[]
    \centering
    \subfloat[Sample of \ac{LDA} topics, and their respective highest probability labels in descending order]{\footnotesize
    \begin{tabular}{|p{0.85in}|p{0.85in}|p{0.8in}|p{0.85in}|p{0.75in}|}
    \hline
\thead[l]{\textbf{Topic 1:}\\ Wintry Conditions}   & \thead[l]{\textbf{Topic 8:} Nighttime\\ Street Lights} &\thead[l]{ \textbf{Topic 9:}\\ Intersection} & \thead[l]{\textbf{Topic 11:} Traffic\\ Congestion }          & \textbf{Topic 12:}  Error     \\ \hline
LS1: snow                   &  LS2: Street                      &  LS2: Intersection       & LS1: vehicle              & LS1: white       \\
 LS2: Snow                       & LS1: street light            & LS1: intersection   &  LS2: Vehicle                  & LS1: material    \\
 LS2: Phenomenon                 &  LS2: Lighting                    & LS1: skyway         & LS1: motor vehicle        &  LS2: Webcam          \\
\makecell[l]{LS1: geological\\\quad phenomenon}  &  \makecell[l]{LS2: Street light }               & LS1: urban area     &  \makecell[l]{LS2: Motor \\\quad vehicle}            & LS1: circle      \\
LS1: phenomenon             & LS1: night                   &  LS2: Urban area         & \makecell[l]{LS1: automotive\\ \quad exterior}  & LS1: technology\\ \hline
\end{tabular}
    \label{tab:selected_topics}} \hspace{2em}
\subfloat[Error message that is shown when a live feed for a camera is unavailable]{
    \raisebox{-0.25in}{\includegraphics[width=.2\linewidth]{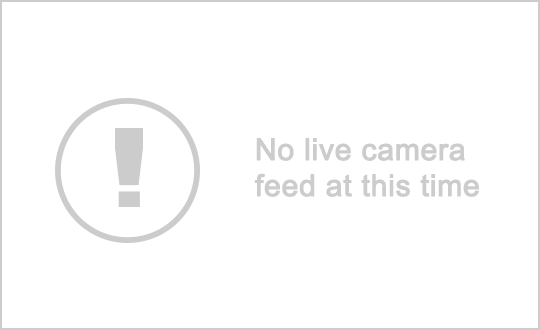}}\label{fig:no_feed}}
    \caption{Selected \ac{LDA} topics \protect\subref{tab:selected_topics} and unavailable feed error message \protect\subref{fig:no_feed}} 
\end{figure*}

\begin{figure}[htb]
    \centering
    \subfloat[Week of Christmas (Mon)]{\includegraphics[width=\linewidth]{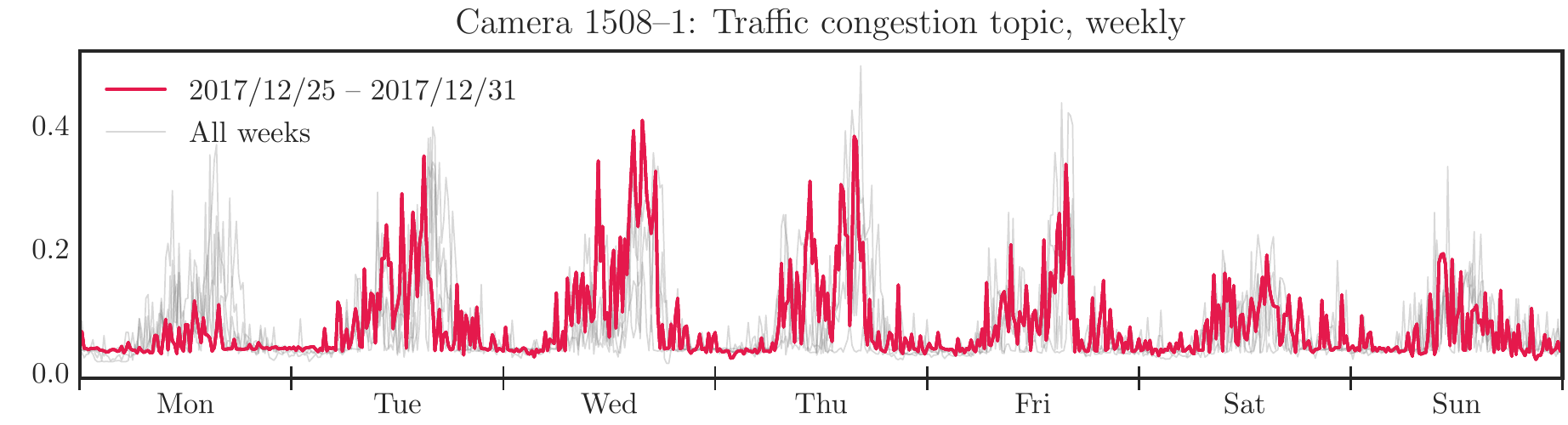}\label{fig:congestion_wk52}}\\
    \subfloat[Week of New Year's (Mon) and ``Bomb cyclone" (Thu--Fri)]{\includegraphics[width=\linewidth]{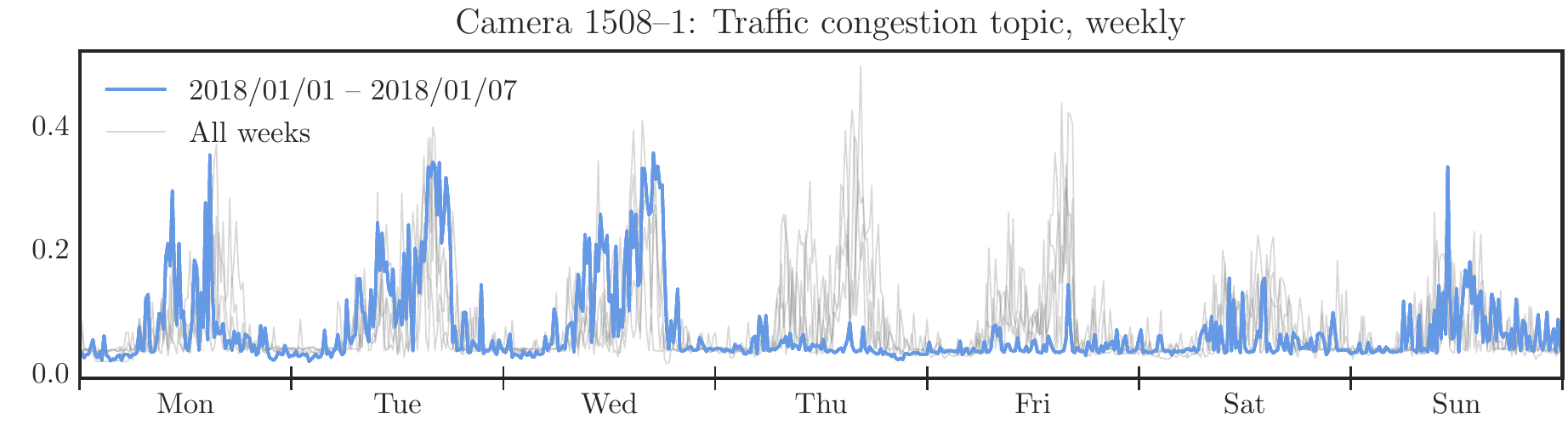}\label{fig:congestion_wk1}}
    \caption{Camera 1508--1 ``traffic congestion" topic signals, with weeks superimposed. Fig.~\protect\subref{fig:congestion_wk52} highlights the week of Christmas; Fig.~\protect\subref{fig:congestion_wk1} highlights the week of New Year's and the ``Bomb cyclone" storm.}
    \label{fig:congestion_topic_ts}
\end{figure}

\begin{figure}[tb]
    \centering
    \subfloat[``LS2: Traffic'' label]{\includegraphics[width=1\linewidth]{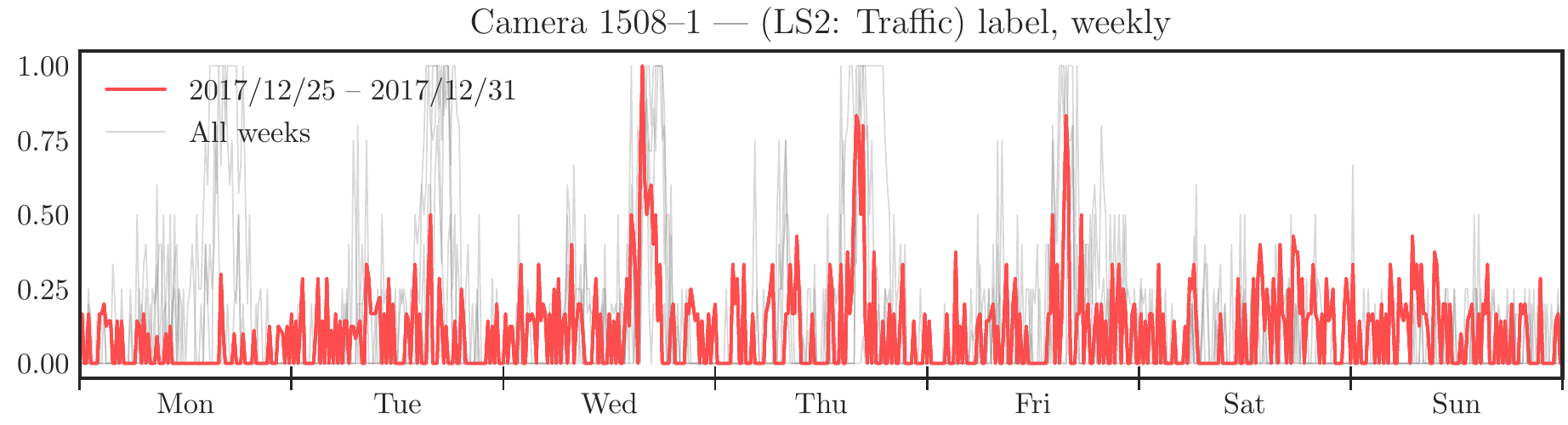}\label{fig:LS2_Traffic}}\\
    \subfloat[``LS2: Traffic congestion'' label]{\includegraphics[width=1\linewidth]{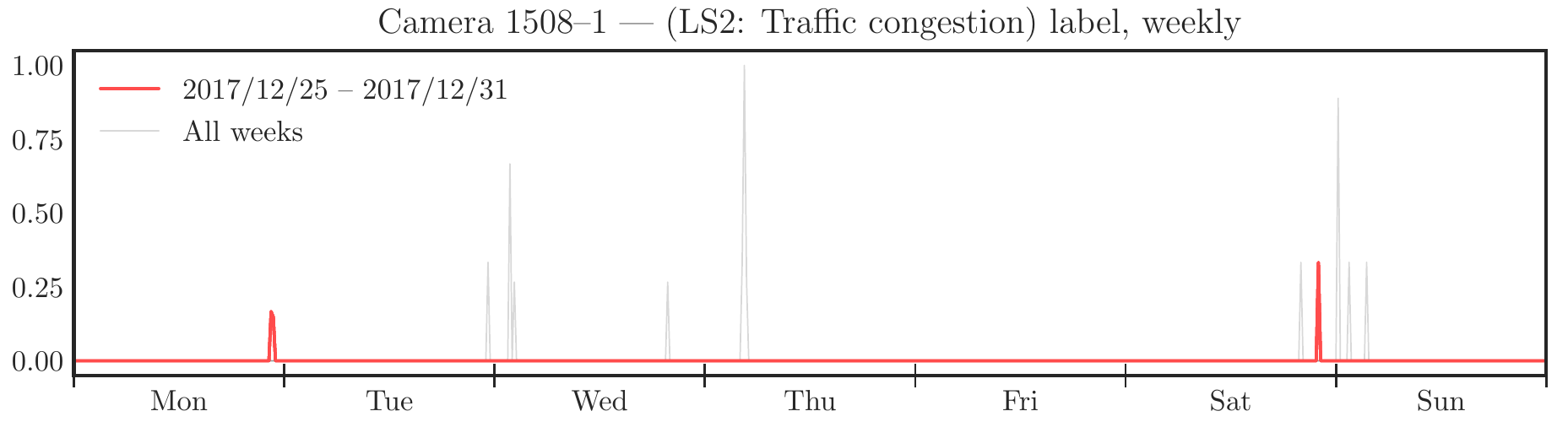}\label{fig:LS2_Traffic_congestion}}
    \caption{Camera 1508--1 individual label signals vs. time, with weeks superimposed. Fig.~\protect\subref{fig:LS2_Traffic} shows the signal for the label ``LS2: Traffic''; Fig.~\protect\subref{fig:LS2_Traffic_congestion} shows the signal for the label ``LS2: Traffic congestion''.}
    \label{fig:traffic_labels_ts}
\end{figure}

We now highlight few selected semantic topics and topic signals. These results come from an \ac{LDA} model with $\numtopicsstar=20$ topics\footnote{Appendix~\ref{sub:num_topics} explains the process for selecting the appropriate number of topics.}, fit on the entire dataset, weighted using the per-camera \ac{tf-idf} scheme \eqref{eq:per_cam_tfidf}. Fig.~\ref{tab:selected_topics} presents a handful of representative topics and their five highest-probability labels. Recall that the \ac{tf-idf} scheme reweights labels relative to their average appearance frequency. Without this reweighting, the highest probability labels of each topic would be dominated by the most common (but less informative) labels of ``road'' and ``asphalt''.

The \ac{LDA} model represents each image in the dataset as a mixture of the $20$ \ac{LDA} topics, where the fraction of each topic corresponds to the fraction of the image's semantic weight associated with that topic. The \emph{semantic topic signal} represents that fraction, for a given topic and camera, as a function of time. We adopt the common practice of naming the topics, \emph{a posteriori}, based on domain knowledge and understanding of the labels in each topic. We associate topics with \emph{processes} which generate labels corresponding to elements related to that process. We find that the topics cover categories of processes including environmental/weather phenomena, diurnal cycles, infrastructure elements, traffic, and error messages.   

It is interesting to note that the semantic meanings are not considered in the \ac{LDA} topic inference process, yet the statistical \ac{LDA} process seems to aggregate semantically similar labels into topics. The exception here is in ``Topic 12: Error," which has labels that are seemingly unrelated to traffic \ac{CCTV} footage, as well as to one another. However, once we realize that Topic 12 appears only for the image shown in Fig.~\ref{fig:no_feed}, which is given by the Mass511 web server when the feed is temporarily down, the relation becomes clear. This semantic similarity and ease of interpretation is an intended feature of the topic model approach, which aims to retain the intuitive parsability of image data. Furthermore, this demonstrates a useful side effect of using the \ac{LDA} representation: automatic identification of frames with recurring error messages.

We find that ``Topic 1: Wintry Conditions'' corresponds to winter storm events. Unsurprisingly, the top labels include ``snow'' from both label sources. However, unexpectedly, the next three labels are variations of ``phenomenon'' and ``geological phenomenon'', which we did not expect \emph{a priori}, to correspond to winter storm events. We find in notable event detection (presented in the next section), that the topic performs better in validation than using naively only ``snow'' and/or ``rain'' labels, suggesting that the ``phenomenon'' and ``geological phenomenon'' labels provide useful information toward detecting winter storm events. This demonstrates another benefit of the semantic topic representation: the ability to discover and identify labels that are related to quantities of interest, and grouping those labels into the same topic.

We now examine the ``Topic 11: Traffic congestion'' topic signal to qualitatively gauge traffic congestion patterns from the footage. Fig.~\ref{fig:congestion_topic_ts} presents the ``Traffic congestion" topic signal for camera 1508--1, with the data from every week superimposed. The data points are plotted at 15-minute intervals (downsampled using mean value). Camera 1508--1 is selected for its location in an underpass, which protects it from atmospheric occlusion due to rain or snow. In addition, for the duration of the data collection period, the camera angle was not manipulated by operators. We observe a diurnal pattern of more traffic congestion during the day, as well as a weekly pattern of lower congestion on the weekends. Furthermore, we see more congestion during the evening rush hours than in the morning, as is expected from typical urban commuting patterns, since the camera is located on a ramp leading out of Boston. 

We also highlight two weeks to show the ``Traffic congestion'' topic signal's sensitivity to holidays and major storms. Fig.~\ref{fig:congestion_wk52} highlights the week of Christmas, which shows a clear reduction in traffic congestion on Christmas day. Fig.~\ref{fig:congestion_wk1} highlights the week of New Year's and the ``Bomb cyclone" winter storm, which occurred on Monday and Thursday--Friday respectively. We see that the New Year's reduction in traffic was not as dramatic compared to that of Christmas; this is consistent with expectations, as in the United States, nearly all businesses and organizations are closed on Christmas, but many businesses are open on New Year's, albeit often with reduced hours \cite{MassLive2018}. 

While the effect of New Year's was relatively mild, the ``Bomb cyclone" had a much more stark impact on traffic, reducing it to effectively zero for much of Thursday and Friday. Though the storm itself did not reach Boston until just past midnight on Friday morning, traffic was virtually nonexistent for all of Thursday. This is likely due to the City of Boston imposing a parking ban, which was in place from 7 a.m. Thursday--5 p.m. Friday \cite{CityOfBostonTweet2018,Andersen}. We see a small uptick in the signal around 5 p.m. on Friday at the end of the ban.   

For comparison, we provide similar weekly graphs constructed using individual labels in Fig.~\ref{fig:traffic_labels_ts}. Like with the topic signal, we plot the label signals at a 15-minute interval using mean-value downsampling. Fig.~\ref{fig:LS2_Traffic} shows the ``LS2: Traffic'' label. We observe that there is a significant background level of noise, with roughly one in every five images being tagged with ``LS2: Traffic'' throughout, including at night. The label signal does seem to capture the general phenomena: the afternoon peak in traffic, and reduced traffic on weekends and holidays. However, it is difficult to distinguish between the smaller variations in this signal, such as differences between weekend daytime and nighttime traffic. Fig.~\ref{fig:LS2_Traffic_congestion} shows the label signal for ``LS2: Traffic congestion''. It is clear that this label signal fails to consistently detect traffic congestion, since the label appears only a handful of times, and generally outside of high-traffic rush hours. The label signal does not capture any of the expected diurnal, weekly, or event-related patterns. The plots for ``LS1: traffic'' and ``LS1: traffic congestion'' were omitted, since the former looks similar to its LS2 counterpart, and the latter does not appear at all on any images for camera 1508--1. 

These graphs suggest that the topic signal provides a better representation of a ``traffic congestion'' process which captures more of the phenomena we expect to observe than label signals do. The following section validates this by comparing the performance of using topic signals versus label signals for detecting notable events.


\section{Identifying Notable Events}
\label{sec:anomaly_detection}

In this section, we address the detection of notable events from topic signals. 
We consider two classes of ``notable'' events. First, we address detecting changes in processes that are nominally stationary: for example, nominal weather that is briefly interrupted by storms. Second, we address detecting anomalies in processes that are non-stationary, but have regular temporal patterns and distributions, such as traffic congestion.

Events of the first class are detected using change-point detection. We demonstrate this in Sec.~\ref{sub:detecting_weather_events} by detecting changes in the mean value of the ``Wintry Conditions'' topic signal to identify inclement weather events. Events of the second type are detected using anomaly detection for samples of data. We demonstrate in Sec.~\ref{sub:anomalous_traffic} the detection of anomalous traffic patterns from the ``Traffic congestion'' topic signal. We validate the performance against known winter storms, holidays, and events. 

Furthermore, we examine the performance of using our topic signal representation versus using individual label signals. In particular: we evaluate the performance of the label signals for ``blizzard", ``rain" and ``snow" to serve as benchmarks in the winter storm detection task. 
For the task of detecting anomalous traffic congestion, we compare to the performance of using the label signals of ``traffic", ``traffic congestion", ``car" and ``vehicle."

For generality, we will use the notation $\dataset$ to refer to a set of data. We will consider sets of data constructed from both topic signals $\topicsignal$ \eqref{eq:topic_signal} as well as label signals $\labelsignal$ \eqref{eq:label_signal}. We compare the performance of both signal types for detecting known notable events in our empirical validation, and demonstrate that . 

\subsection{Detecting Changes in Stationary Processes: Winter Storms} 
\label{sub:detecting_weather_events}

\begin{figure}[h!]
\centering
\subfloat[Camera 1106--1 Topic Signal]{\includegraphics[width=0.5\linewidth]{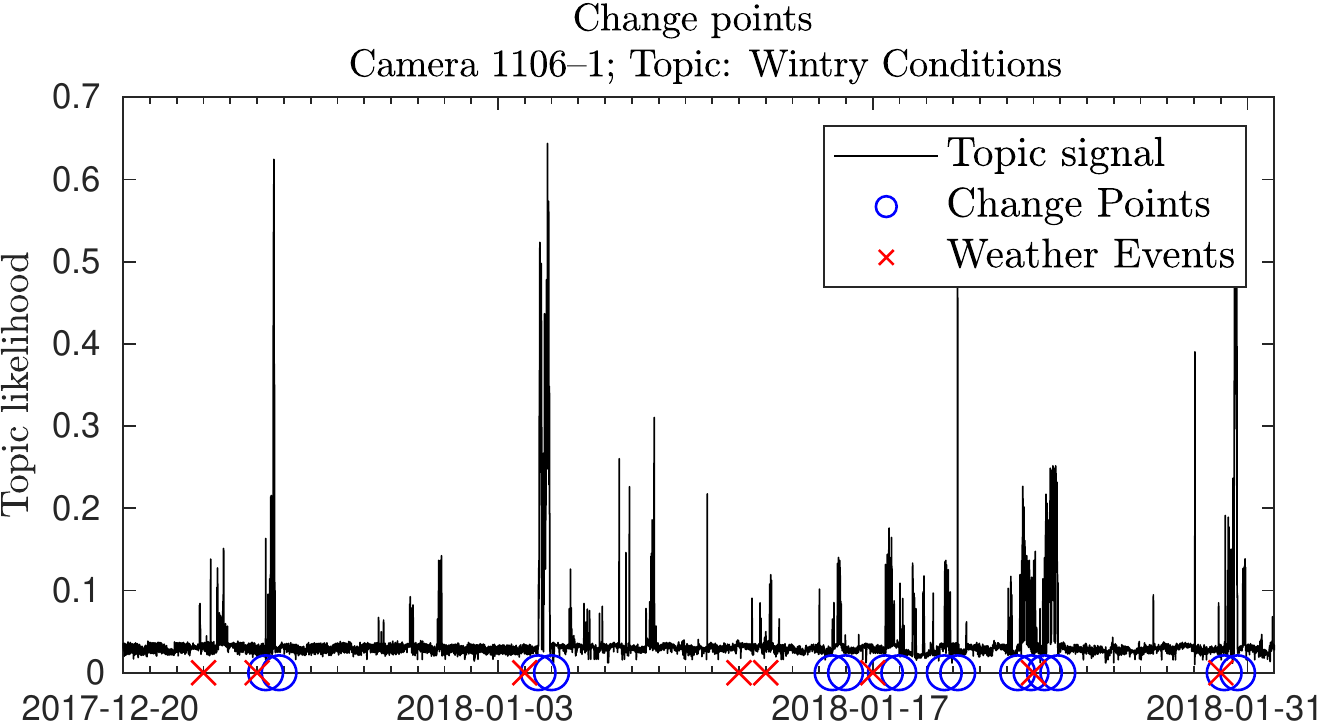}\label{fig:topic_sig_changept1}}
\subfloat[Camera 1137--1 Topic Signal]{
\includegraphics[width=0.5\linewidth]{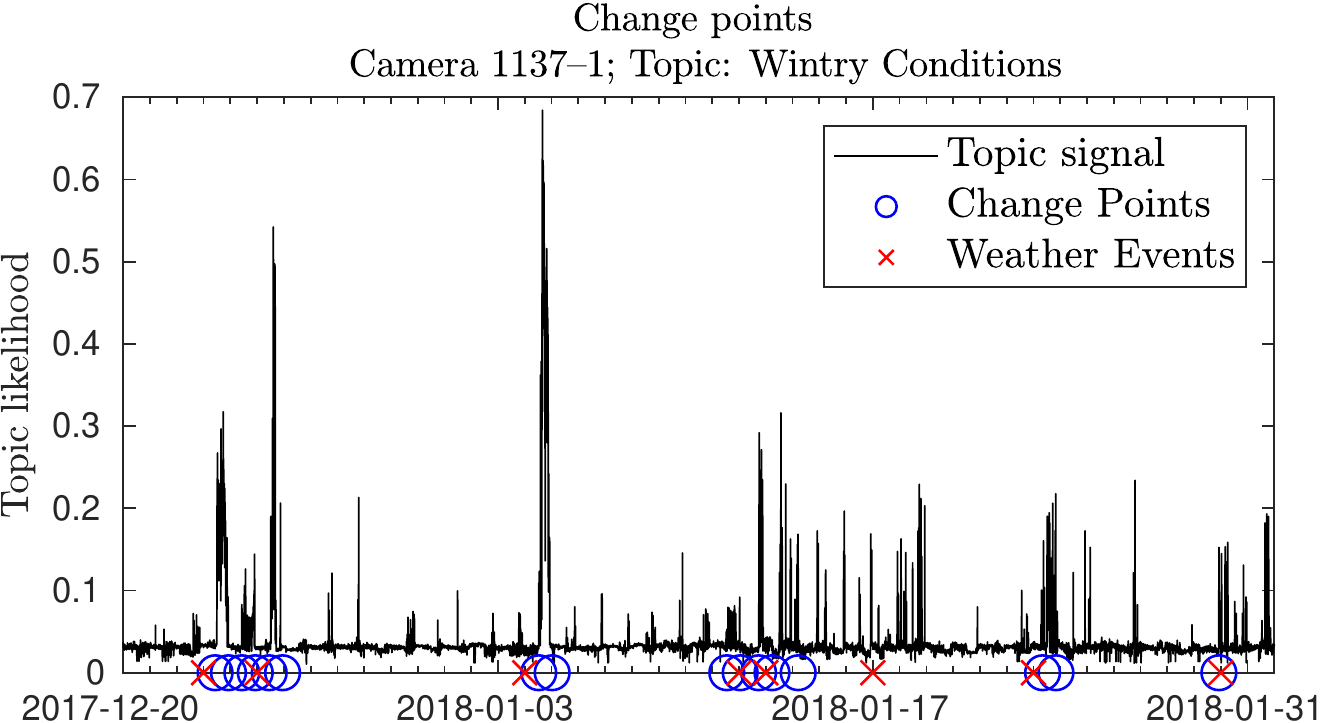} \label{fig:topic_sig_changept2}}\\
\subfloat[Performance evaluation for change-point detection using label signals compared to topics signal. Best scores in each column are rendered in \textbf{bold}]{\footnotesize\begin{tabular}{l|ll|l||ll|l}
    \textbf{Camera}                         & 1106--1    &                &                & 1137--1    &                &                \\ \hline
                         & Prec          & Rec              & $F_1$          & Prec          & Rec              & $F_1$          \\\hline
LS1: snow                & 0.5        & 0.5            & 0.5            & 0.5        & 0.5            & 0.5            \\
LS2: Snow                & 0.5        & 0.5            & 0.5            & 0.75       & 0.375          & 0.5            \\
LS1: blizzard            & \textbf{1} & 0.375          & 0.5455         & \textbf{1} & 0.25           & 0.4            \\
LS2: Blizzard            & \textbf{1} & 0.375          & 0.5455         & \textbf{1} & 0.25           & 0.4            \\
LS1: rain                & \textbf{1} & 0.375          & 0.5455         & 0.857      & 0.75           & 0.8            \\
LS2: Rain                & \textbf{1} & 0.375          & 0.5455         & 0.857      & 0.75           & 0.8            \\ \hline
\makecell[l]{``LS1: rain \\\quad OR \\LS1: snow''}   & 0.5        & 0.5            & 0.5            & 0.875      & \textbf{0.875} & \textbf{0.875} \\\hline
\makecell[l]{Topic:\\ ``Wintry\\ Conditions''} & 0.625      & \textbf{0.625} & \textbf{0.625} & 0.875      & \textbf{0.875} & \textbf{0.875}
\end{tabular}\label{tab:perf_compare_changept}}
    \caption{Performance of winter storm detection using change point detection. Figures in~\protect\subref{fig:topic_sig_changept1}--\protect\subref{fig:topic_sig_changept2} show the change points in the ``Wintry Conditions'' topic signal and weather events for cameras 1106--1 and 1137--1. Table~\protect\subref{tab:perf_compare_changept} compares the performance of using the topic signals to using label signals. }
    \label{fig:changept}
\end{figure}

We first consider detecting deviations from stationary processes; that is, processes which typically have a constant mean and variance, but are occasionally disturbed by transitory disruptions. Winter storm disruptions to nominal weather conditions can be modeled as such a process. Under normal weather conditions, a measurement of a ``winter storm" process should be constant at zero; however, whenever there is a storm, that process should have a positive, nonzero measurement. If a signal captures this behavior, then we can identify notable events by detecting changes in the mean of that signal.  

We use change point detection to identify the notable events. Change point detection is the problem of finding points in time series where the statistics of the data on either side differ significantly \cite{lavielle_using_2005}. In our case, we are looking for the points in time where the mean of the preceding and subsequent data differ significantly. This can be posed as an optimization problem \cite{lavielle_using_2005} of finding the vector $\changeptvec$ of $\numchangepts$ change points which minimizes the following objective function:
\begin{align}
\sum_{\changeptidx=1}^{\numchangepts}[\mathcal{C}(\dataset_{\changeptvec_{\changeptidx-1}:\changeptvec_\changeptidx}) + B] \label{eq:change_pt_obj},
\end{align}
where $\changeptvec_\changeptidx$ denotes the $\changeptidx^\text{th}$ changepoint; $\dataset_{\changeptvec_{\changeptidx-1}:\changeptvec_\changeptidx}$ denotes the data points of dataset $\dataset$ that fall between the change points $\changeptvec_{\changeptidx-1}$ and $\changeptvec_\changeptidx$; and $B$ is a constant parameter to prevent overfitting. The elements of the change point vector are sequentially ordered in time, i.e. $\changeptvec_\changeptidx < \changeptvec_{\changeptidx'}$ iff $\changeptidx < \changeptidx'$.
The cost function is defined as $\mathcal{C}(\xi) = \|\xi- \mu_\xi \|_2$, the $L_2$ norm of the difference between a subsample of data $\xi$ from its mean $\mu_\xi$.

Essentially, minimizing \eqref{eq:change_pt_obj} finds the change points that result in the best fit of a piecewise constant signal to the data. The parameter $B$ prevents overfitting by acting as a minimum threshold: an additional change point is only added if it can reduce the sum of squared difference from the means by at least $B$. We choose $B$ to be $1/20$ of the total energy ($L_2$ norm) of the original signal: $B = \|\dataset\|_2/20$.

We use this technique to find the changes in the ``Wintry Conditions" topic signals from cameras 1106--1 and 1137--1, presented in Fig.~\ref{fig:topic_sig_changept1} and \ref{fig:topic_sig_changept2}, respectively; i.e. $\dataset = \topicsignal$ for $\topic=$``Wintry Conditions'' and $\camera=$1106--1, and 1137--1. The change points are validated against the eight events in the ``Rain'' and ``Snow'' columns in Table~\ref{tab:anomalies}. Since the frequency of the weather data provided by \ac{NOAA-GHCN} is daily, we allow for a $\pm 12$ hour detection window around the change points---i.e. if a weather event happens within a $12$ hour window of a detected change point, it is a true positive. This accounts for the temporal uncertainty due to the 24-hour quantization of the reported weather data. Furthermore, we consider pairs of change points as a single detection event: the first change point represents the start of the event (deviation from nominal), and the second represents the end (return to nominal). We also consider additional change points which happen within 24 hours of a start of a detection event as part of the same event, and thus are not counted as additional change points. This 24-hour minimum duration was chosen to match the \ac{NOAA-GHCN} data. Finally, we consider (up to) the top eight significant detection events for each signal. 

We evaluate the performance of the event detector using the classical $F_1$ score, which is the geometric mean between the precision (Prec), and recall (Rec) metrics \cite{Manning2009a}. Precision measures the fraction of positive classifications which are correct, and recall measures the fraction of total events which are detected. They are given as:
$F_1 = \frac{\text{Prec}\cdot\text{Rec}}{\text{Prec}+\text{Rec}},$ where
    $\text{Prec} = \frac{\text{TP}}{\text{TP}+\text{FP}}$; $\text{Rec} = \frac{\text{TP}}{\text{TP}+\text{FN}};$
TP stands for True Positives; FP for False Positives; and FN for False Negatives. 

Fig.~\ref{tab:perf_compare_changept} presents the performance metrics of the changepoint detection applied to the ``Wintry Conditions'' topic signal and compares it against the use of various label signals.
We evaluate the performance of the label signals for ``snow,'' ``rain,'' and ``blizzard'' from both LS1 and LS2, as well as all pairwise combinations of those labels. We show only the best performing pairwise combination: ``LS1: rain OR LS1: snow.'' We see that in all cases, as measured by $F_1$ score, the detection events from the topic signal outperform those from the label signals. In addition, it performs as well as, or better than, the performance of the best pairwise combination of labels. While the label signals of ``blizzard'' and ``rain'' achieved higher precision, their recall, and thus $F_1$ score, was much worse: i.e. they correctly identified a small number of events, but completely missed the rest.


\subsection{Detecting Anomalies in Non-Stationary Processes: Traffic}
\label{sub:anomalous_traffic}

Certain processes are inherently non-stationary; for example, the traffic congestion process follows a diurnal pattern of increasing during morning and evening rush hours, and decreasing to zero at night. This non-stationarity prevents us from using the previously discussed change point detection approach to detect notable events. One way to address this would be to model traffic congestion as a trend-stationary process: i.e. a sum of a deterministic time-dependent diurnal trend component and a stationary stochastic component and detect changes in the stochastic component. However, this requires the estimation of the trend component, which introduces another modeling and statistical question. 

Instead, we present an alternative approach which sidesteps the need to estimate the temporal trend signal. We pose the problem as a statistical anomaly detection problem by measuring the dissimilarity (via an $f$-divergence measure) between the empirical distribution of the signal values and a set of nominal reference distributions. Furthermore, we employ a direct estimation technique \cite{Sugiyama2007,Yamada2013} for computing the divergence between two empirical distributions without having to parametrically estimate the distributions themselves as well. Our method offers significant generality, as it does not depend on the functional forms of the temporal trend or distribution.

In addition, we consider data \emph{subsequences} as our ``data points."  A subsequence starting at data point $\datapoint_\imageidx \in \dataset$ is represented as
$\chi_\imageidx = [\datapoint_{\imageidx},\datapoint_{\imageidx+1},\dots, \datapoint_{\imageidx+k-1}] \in \mathbb{R}^{mk},$
where $k$ is the subsequence length and $m$ is the number of dimensions of $\datapoint_\imageidx$. We then use the set of subsequences $\boldsymbol{\chi} = \{\chi_\imageidx\}_{\imageidx=1}^{\numimages-k}$ as the dataset for anomaly detection, where $N$ is the number of elements of $\dataset$. This process is similar to the construction of the lag terms in autoregressive (AR) models, and is used in other time series analysis problems \cite{Liu2013b}. We vary the length subsequence length $k$ and empirically determine the best subsequence length to use in our anomaly detection procedure. 

Our approach considers the anomaly detection problem of whether a test sample of data from a signal is \emph{anomalous} compared to a set of known nominal reference samples. Let us divide the length of a signal into $\numwindows_\dataset$ equal-sized time windows, where $T_s$ denotes $s^\text{th}$ window and $s \in [1,\numwindows_\dataset]$ indexes the windows. Let $\dataset_s$ denote a \emph{test} sample of data corresponding to the data points which occur during $T_s$. Let $\refcorpus_\sigma$ similarly denote a reference sample of nominal data, where $\sigma \in [1, \numwindows_\refcorpus]$ indexes the reference samples, and the set of all reference samples is denoted $\refcorpus$. We determine whether a sample $\dataset_s$ is anomalous based on its average dissimilarity to the reference samples $\refcorpus_\sigma \in \refcorpus$.

\subsubsection{Divergence Measures and Anomaly Detection}
We compute dissimilarity between two data distributions using an $f$-divergence, defined as follows: for probability distributions $P, P^\prime$, defined over a space $\Omega$ (with respective probability densities $p(\boldx), p^\prime(\boldx)$), an $f$-divergence from $P$ to $Q$ is given by:
\begin{align}
    \fd{P||P^\prime} := \int_\Omega p^\prime(\boldx) f\left(\frac{p(\boldx)}{p^\prime(\boldx)}\right) d\boldx \label{eq:f_divergence}
\end{align}
where $f(t)$ is a convex function with $f(1) = 0$ \cite{ali1966general}. The well-known Kullback-Leibler (KL) divergence, and Pearson (PE) $\chi^2$-divergence are specific instances of $f$-divergences, where $f_{KL}(t) = t \log(t)$ and $f_{PE}(t) = \frac{1}{2}(t-1)^2$ respectively \cite{ali1966general}.

All $f$-divergences are positive, are minimized at zero when $P$ and $P^\prime$ are identical, and maximized when they are statistically independent; in addition, they satisfy information monotonicity and joint convexity \cite{ali1966general}. A larger $f$-divergence value indicates a greater dissimilarity between two distributions than a smaller divergence; note however, that $f$-divergences are not true distance measures, in that they are not commutative, i.e. $D_f(P||P^\prime)\neq D_f(P^\prime||P)$, and do not satisfy the triangle inequality. In our application, we adopt the common practice \cite{Liu2013b} of using a symmetrized divergence which satisfies commutativity, given as
$\symfd{P||P^\prime}  := \fd{P||P^\prime} + \fd{P^\prime||P}\nonumber$.

In this paper, we consider a variant of the PE divergence, the Relative Pearson (RP) divergence \cite{Yamada2013}, defined as:
\begin{align}
    \rpd{P||P^\prime} = \frac{1}{2}\int_\Omega q_\gamma(\boldx) \left(\frac{p(\boldx)}{q_\gamma(\boldx)} - 1\right)^2 d\boldx \label{eq:RP_divergence}
\end{align}
where 
\begin{align}
    q_\gamma(\boldx) = \gamma p(\boldx) + (1-\gamma)p^\prime(\boldx), \label{eq:alpha_relative_density}
\end{align}
for some $\gamma \in [0,1)$, is referred to as the $\gamma$-relative density \cite{Yamada2013}.\footnote{The notation in \cite{Yamada2013} refers to this quantity as the $\alpha$-relative density. However, in this paper, we refer to it as the $\gamma$-relative density to avoid the ambiguity with the $\alpha$ \ac{LDA} hyperparameter.} The use of $q_\gamma(\boldx)$ in \eqref{eq:RP_divergence} ensures that the $\gamma$-\emph{relative density ratio}, $r_\gamma(\boldx) = p(\boldx)/q_\gamma(\boldx)$, stays upper bounded by $\frac{1}{\gamma}$. This boundedness improves the rate of numerical convergence when estimating the divergence \cite{Yamada2013}.

We estimate the divergence using the \ac{RuLSIF} direct estimation procedure presented in \cite{Yamada2013}. 
\ac{RuLSIF} is an extension of direct divergence-estimation procedures such as the \ac{KLIEP} for estimating KL divergence \cite{Sugiyama2007} and \ac{uLSIF} for estimating the Pearson divergence \cite{Kanamori2009}. These direct estimation procedures estimate the divergence measure between two sets of data without the need to parametrically estimate the respective distributions $p(\boldx)$ and $p^\prime(\boldx)$ of each data set. This is quicker to compute and more accurate in estimating divergences than estimating the distributions separately \cite{Sugiyama2007, Kanamori2009, Yamada2013}. We choose \ac{RuLSIF} in particular because it computes quicker when compared to the similar \ac{uLSIF} and \ac{KLIEP} techniques \cite{Yamada2013}.  

We construct an anomaly score for a test sample $\dataset_s$ and set of reference samples $\refcorpus$ based on the Relative Pearson divergence. We refer to this anomaly score as the \acf{RPDAS}. It is computed as the average symmetrized RP divergence between the test sample and each of the reference samples, given as:
\begin{align}
    \text{RPDAS}(\dataset_s, \refcorpus) := \frac{1}{\numwindowsprime}\sum_{\sigma=1}^{\numwindowsprime} \symestrpd{\dataset_s, \refcorpus_{\sigma}}, \label{eq:rpdas}
\end{align}
where $\symestrpd{\dataset_s, \refcorpus_{\sigma}}$ is the symmetrized, \ac{RuLSIF}-estimated RP divergence between the test sample $\dataset_s$ and reference sample $\refcorpus_{\sigma}$. The RPDAS is bounded on the same range as the RP divergence: $[0,1/\gamma]$. 

A sample $\dataset_s$ is flagged as a \emph{detection event} if the \ac{RPDAS} of that sample exceeds an alert threshold $\thresh$. An anomaly detection event for sample $\dataset_s$ is considered a true positive detection if there is a true \emph{anomaly event} during the time period $T_s$. False positives correspond to detection events without a corresponding true anomaly event, and false negatives correspond to missed detections of true anomaly events. By varying the threshold $\thresh \in [0, 1/\gamma]$, we can adjust the sensitivity of the anomaly detection process. In this way, we compute a Precision-Recall (PR) curve \cite{Manning2009a} to evaluate performance. We use both the area under the PR curve (PR AUC), as well as the configuration with the best $F_1$ score as performance metrics. The PR AUC evaluates the overall performance of the anomaly detector on a range of $[0,1]$, with 1 being a perfect score \cite{Manning2009a}, while the best $F_1$ score evaluates the best-case performance of the detector.

\subsubsection{Empirical Validation}
\label{ssub:anom_score_results}

We now validate our approach to detecting notable events via anomaly detection on topic signals. We consider the process of traffic congestion, as measured by the ``Traffic congestion'' topic signal. 
We focus on the data from camera 1508--1, which was the only camera in the dataset with no changes in camera angle or perspective. In addition, its location in an underpass protects it from atmospheric interference and obstruction. These properties help ensure that any variations in traffic congestion that we detect reflect the actual changes in in the process, and not caused by external factors.  

We detect days with anomalous distributions of data points and subsequences in the topic signal $\Theta_{1508\text{--}1}^\text{Traffic congestion}$ using the \ac{RPDAS}, and compare these detection events with known anomaly events which we expect to significantly affect traffic congestion. These anomaly events are the ``holiday/special event'' and ``snow'' columns of Table~\ref{tab:anomalies}. We did not consider rain-only events, as we believed they were less likely to cause disruptions to traffic compared to snowfall, holidays, or special events. Our validation results seem to support this hypothesis, as none of the signals showed sensitivity to rain-only events. Similarly to the change detection analysis, the data are partitioned into 24-hour-long windows to match the granularity of the event data in Table~\ref{tab:anomalies}. Data from phase I is used as the reference dataset $\refcorpus$. The reference data spans November 6$^\text{th}$--November 12$^\text{th}$, 2017, contained no significant weather events or holidays.

\begin{figure*}[h!]
    \centering
    \subfloat[]{\includegraphics[height=0.4\textheight]{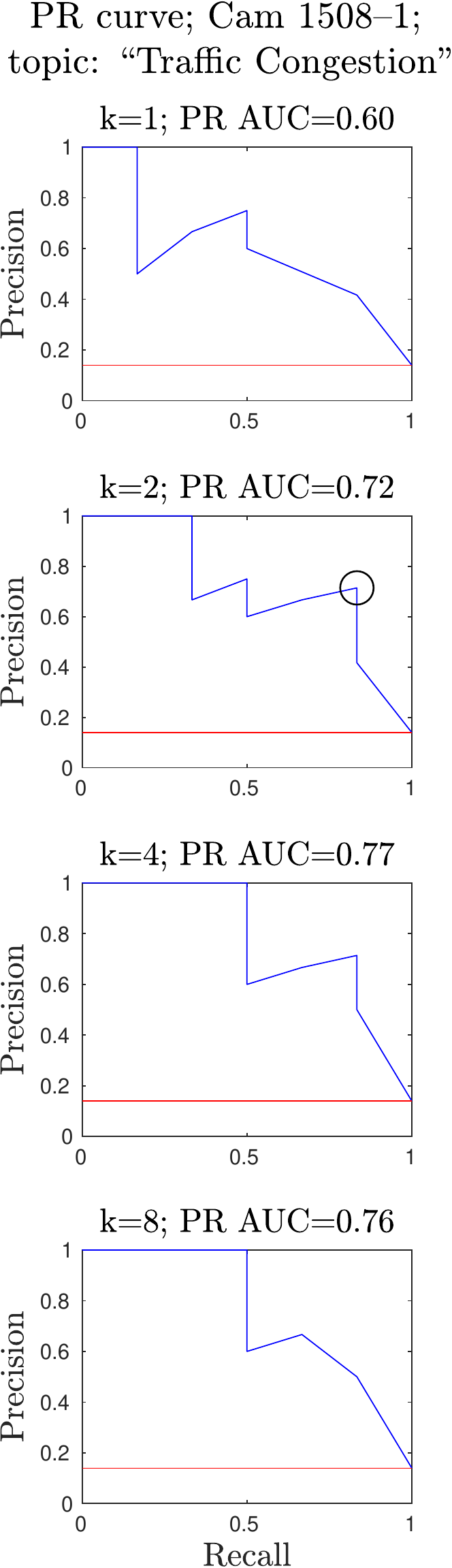}\label{fig:pr_curves_topics}}\hspace{2em} \hspace{0em}
    \subfloat[]{\includegraphics[height=0.4\textheight]{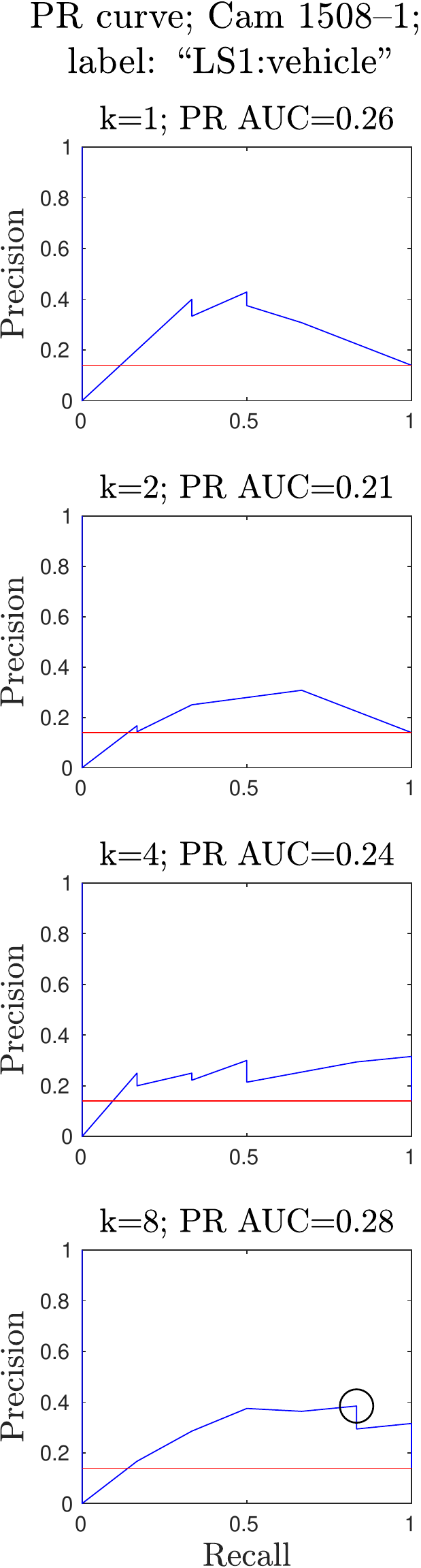}\label{fig:pr_curves_label1}}\hspace{2em}\hspace{-2em}
    \subfloat[]{\includegraphics[height=0.4\textheight]{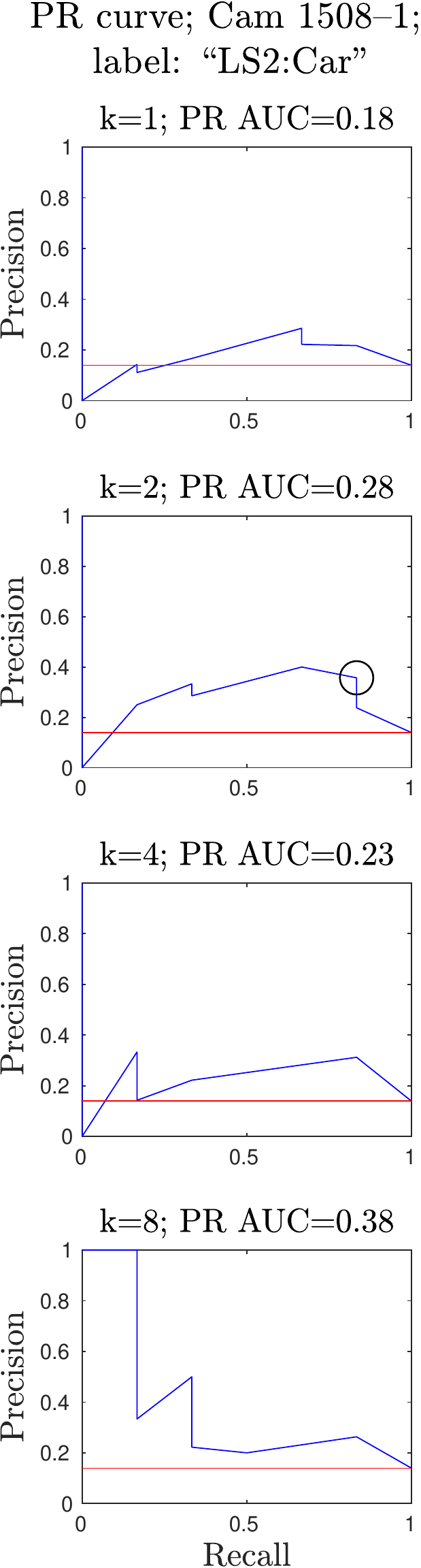}\label{fig:pr_curves_label2}}\hspace{0em}
    \subfloat[]{\includegraphics[height=0.4\textheight]{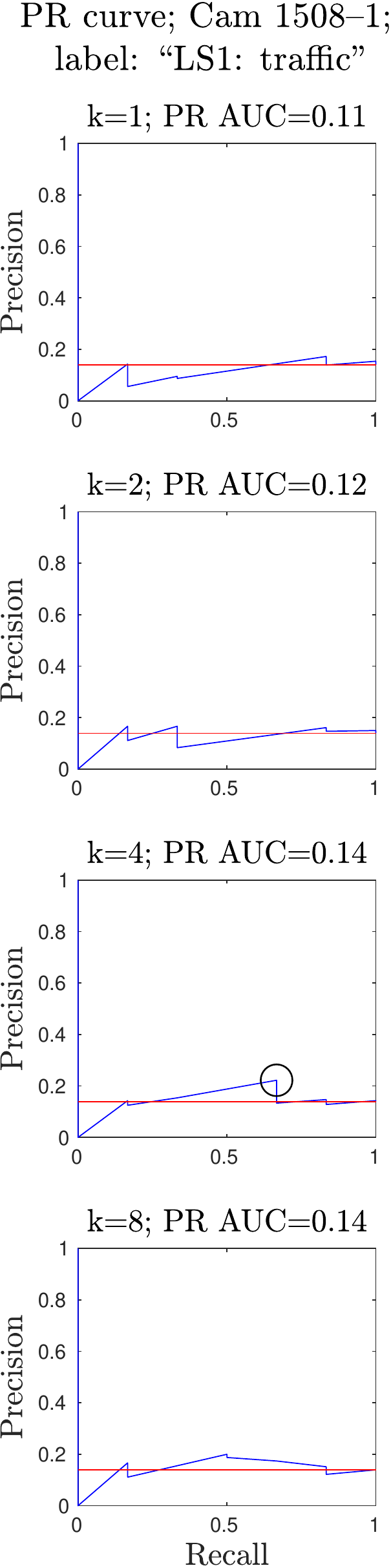}\label{fig:pr_curves_label3}}\hspace{0em}
    \subfloat[]{\includegraphics[height=0.4\textheight]{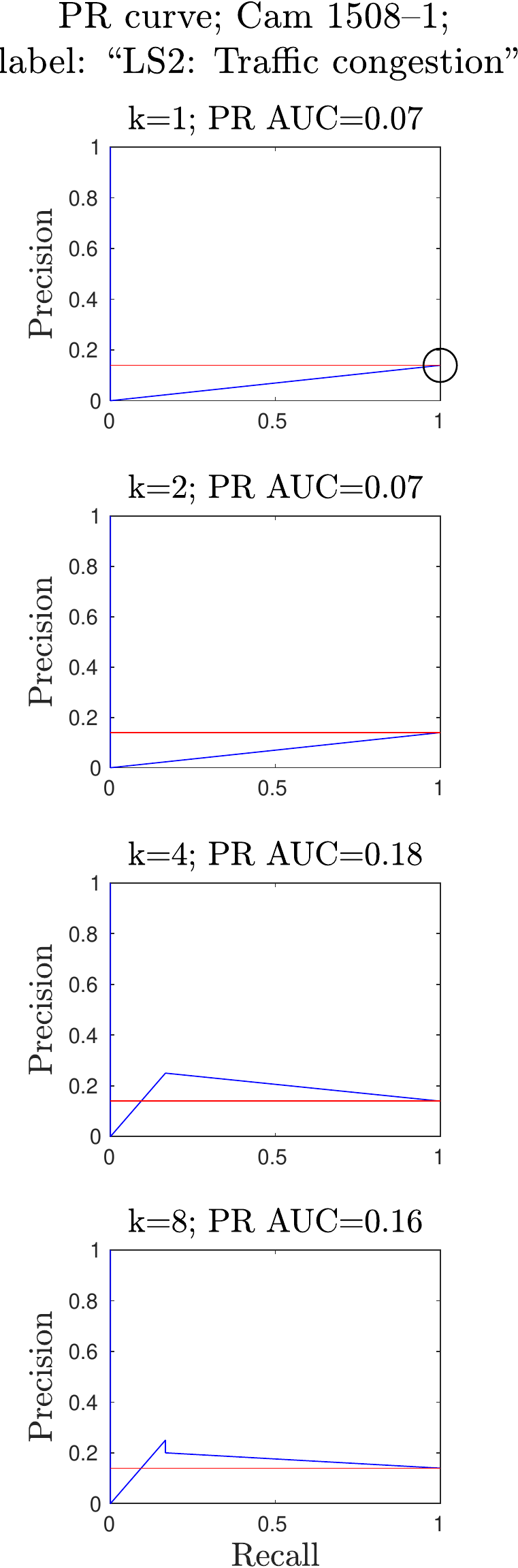}\label{fig:pr_curves_label4}}
    \caption{Precision-recall curves for anomalous traffic detection for various signals and subsequence window lengths $k$. The black circle indicates highest $F_1$ score in each column; the horizontal red line indicates performance of null predictor. Fig.~\protect\subref{fig:pr_curves_topics} shows the results using the \ac{LDA} ``Traffic congestion" topic signal, whereas Figs.~\protect\subref{fig:pr_curves_label1}--\protect\subref{fig:pr_curves_label4} show the results using a number of selected label signals. }
    \label{fig:PR_curve_comparisons}
\end{figure*}

The daily \ac{RPDAS} is computed for $\Theta_{1508\text{--}1}^\text{Traffic congestion}$, with $\gamma=10^{-3}$ for the $\gamma$-relative density parameter, and for various subsequence window lengths $k\in\{1,2,4,8\}$. The threshold $\thresh$ was varied from zero to $1/\gamma$ to construct Precision-Recall curves. The PR curves for all configurations of ($k, \thresh$) are presented in Fig.~\ref{fig:pr_curves_topics}. The null classifier baseline (uniform random guesses) is given by the red horizontal line corresopnding to a PR AUC of 0.14.

\begin{figure}[tb]
    \centering
    \includegraphics[width=\linewidth]{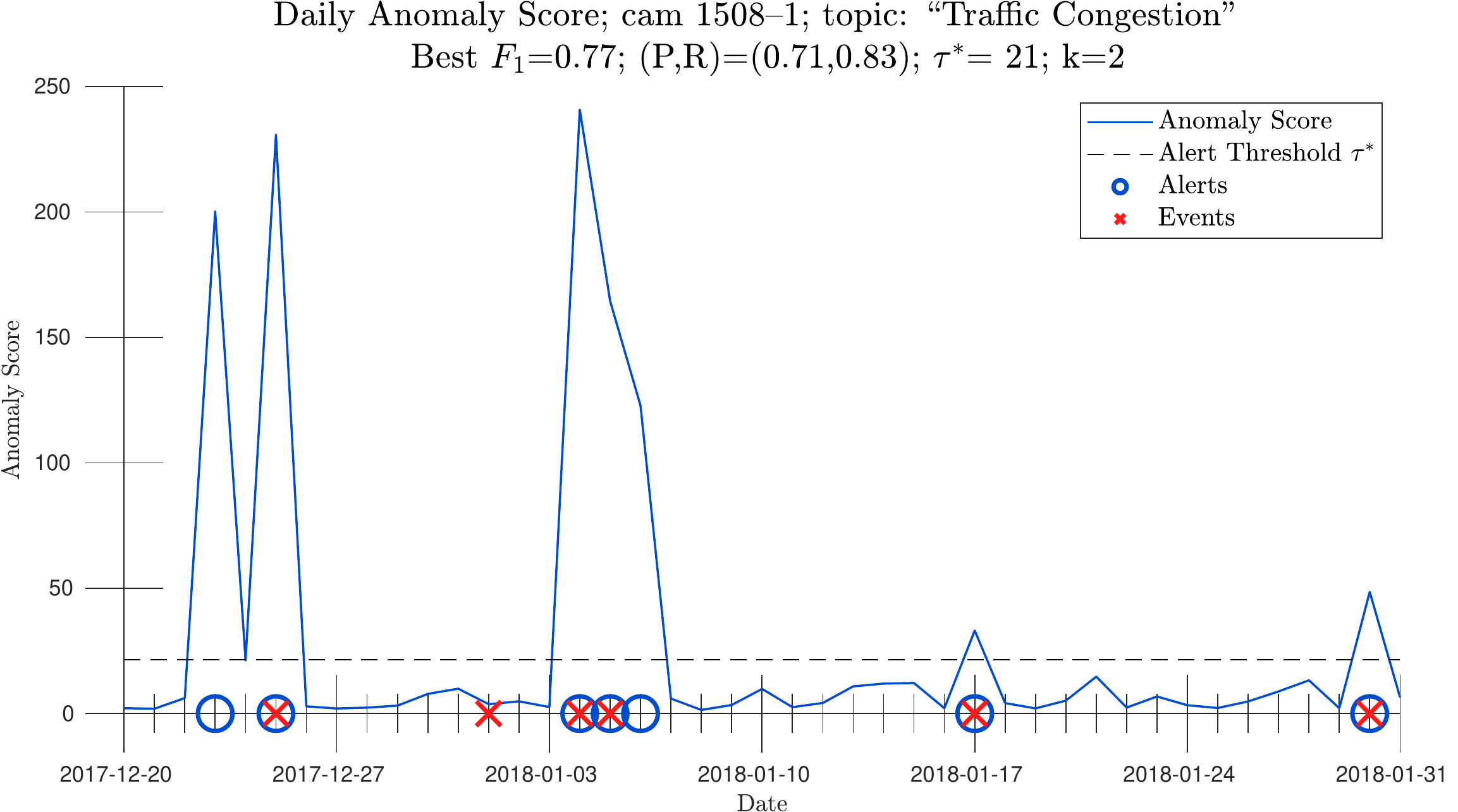}
    \caption{Anomalous traffic detection results for Camera 1508--1. The figure renders the daily anomaly score as a blue line; the alert threshold $\thresh^*$ is depicted as the horizontal dashed line; anomaly events are denoted with a red ``x", and alerts are denoted with a blue ``o". }\label{fig:anom_detect_results}
\end{figure}

Fig.~\ref{fig:anom_detect_results} shows the daily \ac{RPDAS} for the configuration with the best $F_1$ score ($k=2$, $\thresh^*=21$). We see that it performs reasonably well: it has one missed detection of New Year's Day, and two false positives: one the weekend before Christmas, and one right after the bomb cyclone storm. 

For comparison, we also compute the PR curves for various individual label signals which may related to the traffic congestion process, including ``car,'' ``vehicle,'' ``traffic,'' and ``traffic congestion''. Figures~\ref{fig:pr_curves_label1}--\ref{fig:pr_curves_label4} displays the PR curves for each label (for brevity, we only show the better performing label between the two sources). We found that no individual label signal achieved comparable performance in anomaly detection in PR AUC or best $F_1$ score. Furthermore, in cases such as in Fig.~\ref{fig:pr_curves_label2} and~\ref{fig:pr_curves_label2}, anomaly detection on the label signals performs worse than the null classifier. The fact that the topic signal outperforms any individual label signal demonstrates that the performance of the topic signal is not simply due to the performance of its component label signals. Instead, topics capture additional information in the \emph{combinations} of labels, enabling the detection of phenomena beyond what image labeling software is explicitly trained to recognize.   

\section{Discussion and Future Work}
\label{sec:discussion}
Our main contributions in this article are: the \ac{BFCC} dataset of freeway \ac{CCTV} camera footage; the \ac{BoLW} model for representing image contents using semantic features; a novel application of semantic topic modeling to identify and represent processes as semantic topic signals; and a demonstration of using change and anomaly detection on semantic topic signals to identify notable events from traffic \ac{CCTV} footage. This work illustrates the potential for semantics-oriented techniques in analyzing image data. These semantic representations retain much of the intuitive interpretability of images while enabling a lower-dimensional, structured representation.

We emphasize the crux of our approach: analyzing semantic representations of image contents strongly resembles \ac{NLP} problems. We provide the \ac{BoLW} model as a foundational equivalent to the \ac{NLP} \ac{BoW} model. However, we acknowledge that the original \ac{BoW} model is quite dated, and there are now many more sophisticated models for representing semantic features in the \ac{NLP} literature. In particular, concepts such as semantic word embeddings \cite{Bengio2006} can provide vector space representations of semantic features in fewer dimensions than \ac{BoLW}. Likewise, there exist more recent topic models which capture more properties than \ac{LDA}, such as those that model conditional relationships between topics \cite{blei2007correlated,Roberts}. These models may provide more nuanced or sophisticated representations, but this is beyond the scope of this article. The intent of this article is not to claim that our approach is the best for notable event detection. Instead, it is intended as a foundational proof-of-concept which motivates the use of semantic representations of image contents and their analysis using \ac{NLP} techniques.

This paper intentionally uses only textual semantic labels to represent image contents to explore the capabilities of semantics-only representations. In practice, we do not expect that purely-semantic representations will be ideal for most applications (except, perhaps, applications with privacy or bandwidth requirements, which benefit from data de-identification and compression via semantic representations). Instead, we believe that semantic features are complementary to existing data sources. Integrating \ac{BoLW} semantic features to enhance existing \ac{BoVW} computer vision applications to construct multi-modal ``Bag-of-Features'' models, as well as fusing the label and topic signals with other traffic data sources, such as loop detectors and radar, are promising future directions.

Finally, we note the most significant challenge encountered in this paper: the change and anomaly detection struggled with changes in camera perspectives. These changes affect the distributions of image contents, and thus change the distribution of labels and topics in the scene. This triggers change detection, but it could be addressed by reinitializing the change detection whenever the camera angle changes. This also affects anomaly detection, as the reference data are no longer representative. As such, all test samples get flagged as anomalous until the perspective returns to the original view. A possible fix is to maintain separate reference data for each perspective---though, this is only feasible if there are a finite set of possible perspectives. Additional work is required to account for these effects of camera angle changes.

\bibliographystyle{IEEEtranN}

\begin{IEEEbiography}[{\includegraphics[width=1in,height=1.25in,clip,keepaspectratio]{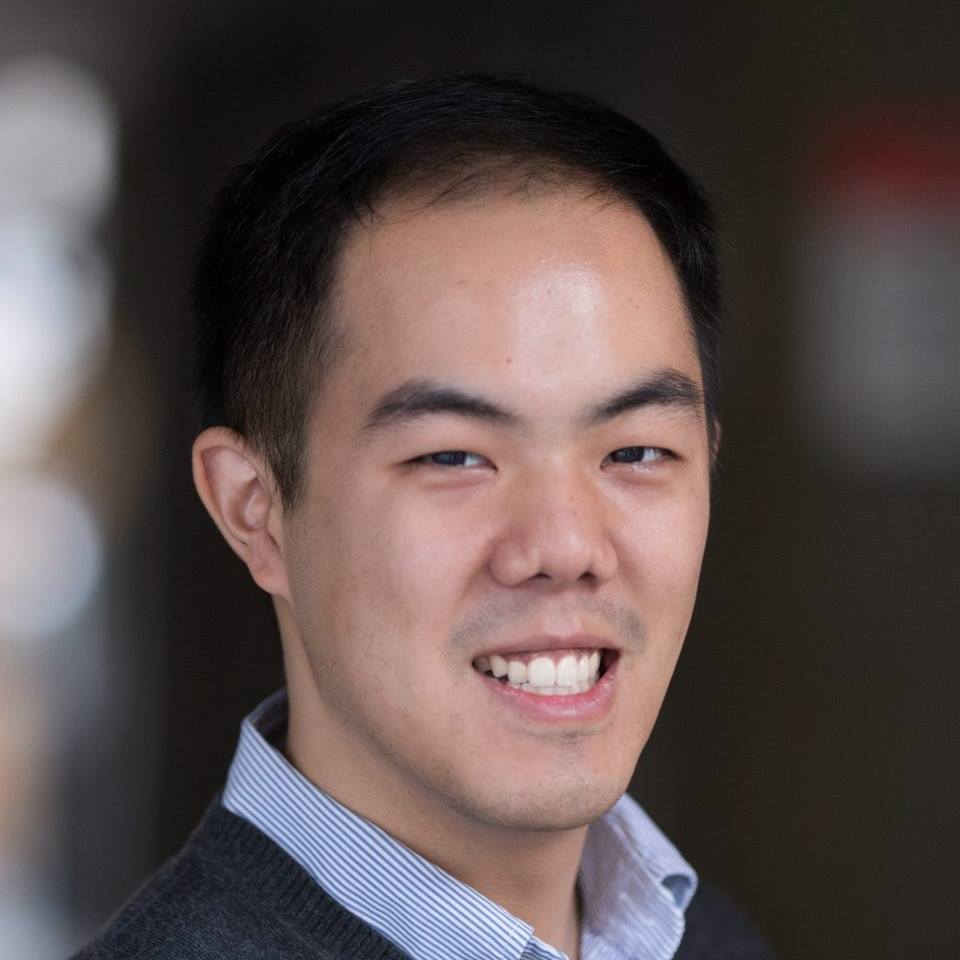}}]{Jeffrey Liu} is a Ph.D. candidate at Massachusetts Institute of Technology in Civil Engineering and Computation, and a member of the Humanitarian Assistance and Disaster Relief Group at MIT Lincoln Laboratory. He received his B.S.E. in Engineering Physics from University of Michigan (2012), and S.M. in Computation for Design and Optimization from MIT (2015). His work addresses disruptions and anomalies in infrastructure networks---focusing on weather disruptions in transportation networks. His research interests include applications of machine learning, computer vision, and natural language processing for public safety and disaster relief.
\end{IEEEbiography}
\begin{IEEEbiography}[{\includegraphics[width=1in,height=1.25in,clip,keepaspectratio]{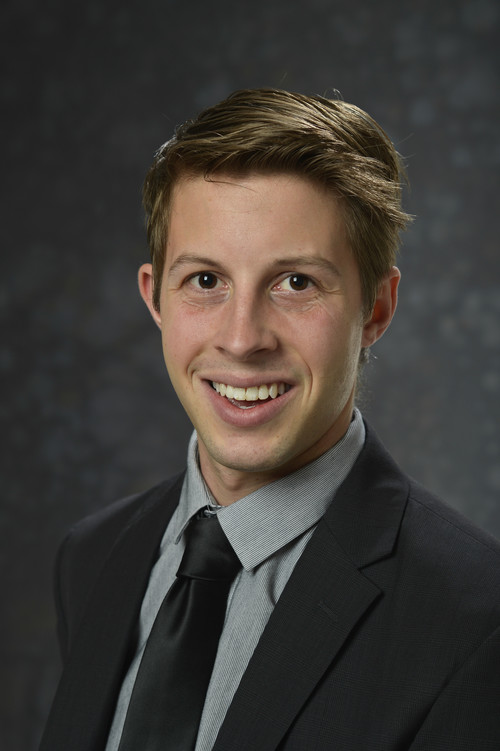}}]{Andrew Weinert} is a member of the Humanitarian Assistance and Disaster Relief Systems Group at MIT Lincoln Laboratory. He received a BS in Security and Risk Analysis with minors in Information Science Technology for Aerospace Engineering and Natural Science from the Pennsylvania State University (2009) and a MS in Electrical and Computer Engineering at Boston University (2014). Mr. Weinert currently serves as the technical lead for a NIST Public Safety Innovation Accelerator Program to generate representative public safety video datasets and leverage edge computing to improve tactical communications.
\end{IEEEbiography}
\begin{IEEEbiography}[{\includegraphics[width=1in,height=1.25in,clip,keepaspectratio]{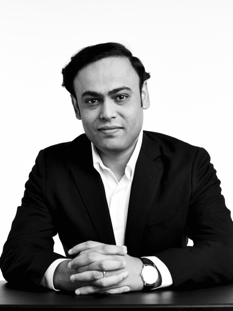}}]
{Saurabh Amin} is a Robert N. Noyce Career Development Associate Professor in the Department of Civil and Environmental Engineering at MIT. He received a B.Tech. (2002) in Civil Engineering from the Indian Institute of Technology at Roorkee, an M.S. (2004) in Transportation Engineering from the University of Texas at Austin, and Ph.D. (2011) in Systems Engineering from the University of California at Berkeley. His research interests are in control of infrastructure networks, cyber-physical systems security, applied game theory and information economics, and optimization in networks.
\end{IEEEbiography}
\pagebreak
\appendices

\section{Choosing Appropriate Number of \ac{LDA} Topics}
\label{sub:num_topics}
The number of topics in the \ac{LDA} model, $\numtopics$, is specified exogenously---i.e. it is not inferred by the model. A larger value of $\numtopics$ can account for more distinct processes, at the expense of increasing model complexity. We use the perplexity metric to choose the appropriate number of topics for the model. Perplexity is an entropy-based metric for assessing how well a probability model predicts an unseen set of test data, $\dataset_\text{test}$ \cite{Blei2003a}, given by:
\begin{align*}
    \text{Perp}(\dataset_\text{test}) := \exp{\left(-\frac{\sum\limits_{\imageidx \in \dataset_\text{test}} \log (p(\labelvec_\imageidx))}{\sum\limits_{\imageidx \in \dataset_\text{test}} \weight_\imageidx} \right)}.
\end{align*}
where $p(\labelvec_\imageidx) $ is the likelihood of the model generating the label vector $\labelvec_\imageidx$. In our case, we use $p(\labelvec_\imageidx) = p(\labelvec_\imageidx|\topicwordparam,\topicworddist)$, the conditional likelihood of observing $\labelvec_\imageidx$ from a \ac{LDA} model given the hyperparameter $\topicwordparam$ and fitted topic-label distribution $\topicworddist$: 
\begin{align*}
    p(\labelvec_\imageidx|\alpha,\topicworddist) = \int p(\doctopicdist_\imageidx|\alpha) \left( \sum_{j=1}^{\overline{\weight}_\imageidx} p(\labelword_j| \topic_j, \topicworddist) p(\topic_j|\doctopicdist_\imageidx)\right) d \doctopicdist_\imageidx.
\end{align*}

\begin{figure}[h!]
    \centering
    \includegraphics[width=0.5\linewidth]{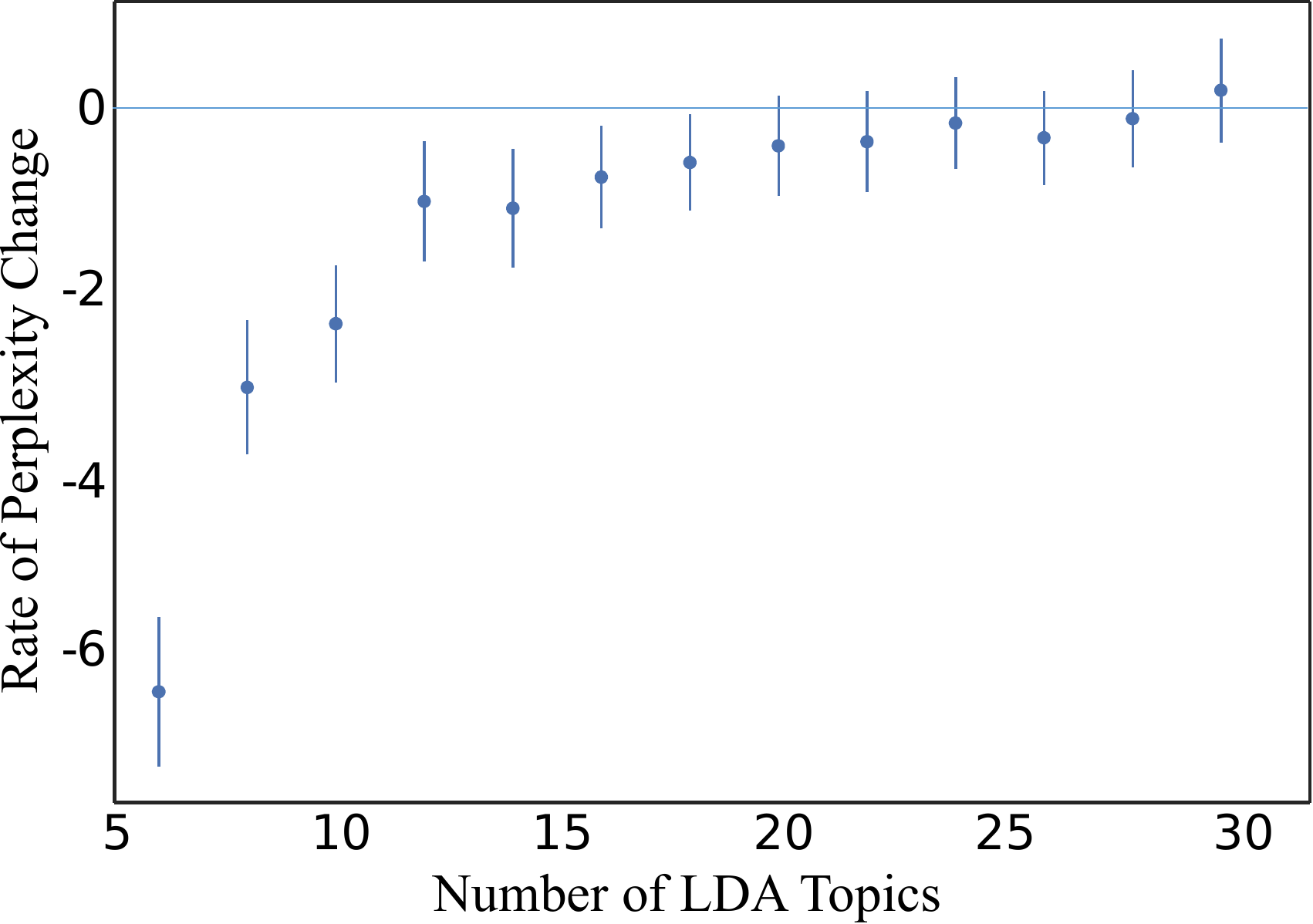}
    \caption{Rate of perplexity change vs. Number of topics; error bars show standard deviation from Monte Carlo samples}
    \label{fig:rpc_err_v_k}
\end{figure}

We select the appropriate number of topics, denoted $\numtopicsstar$, in a manner similar to \cite{Zhao2015}. Since a lower perplexity score indicates a better fit of the model to the data, we increase $\numtopics$ until we no longer see an appreciable decrease in perplexity. 
Let $\text{Perp}_\numtopics(\dataset_\text{test})$ denote the perplexity of a holdout dataset $\dataset_\text{test}$ for an \ac{LDA} model with $\numtopics$ topics. The data was partitioned at random into an $80/20$ train/test split. 
Several \ac{LDA} models were fit over a range of $\numtopics$, and we compute the Rate of Perplexity Change---a finite difference approximation of the slope with respect to $\numtopics$---as:
\begin{align*}
    \text{RPC}(K) := \frac{\text{Perp}_\numtopics (\dataset_\text{test}) - \text{Perp}_{\numtopics-\Delta \numtopics} (\dataset_\text{test})}{\Delta \numtopics}.
\end{align*}
Figure~\ref{fig:rpc_err_v_k} shows the rate of perplexity change versus the number of topics; the error bars represent the standard deviation of 50 Monte Carlo resamplings, with random train/test data partitions for each resampling. We select the smallest $\numtopics$ within one standard deviation from zero as the number of topics, $\numtopicsstar=20$.



%



\end{document}